\documentclass{egpubl}
\usepackage{eurovis2022}

\STAR                   
\STAREurovis   

\usepackage[T1]{fontenc}
\usepackage{dfadobe}  

\usepackage{cite}  
\BibtexOrBiblatex
\electronicVersion
\PrintedOrElectronic
\ifpdf \usepackage[pdftex]{graphicx} \pdfcompresslevel=9
\else \usepackage[dvips]{graphicx} \fi

\usepackage{egweblnk}
\usepackage{color}
\usepackage[dvipsnames]{xcolor}
\usepackage{makecell}
\usepackage{multirow} 
\usepackage{amssymb} 
\usepackage{graphicx} 
\usepackage{array} 
\usepackage{tikz}
\usepackage{collcell}
\usepackage{tabularx,colortbl}
\usepackage{amssymb}
\usepackage{pifont}
\usepackage{subcaption}
\usepackage{soul}

\newcommand{\cmark}{\ding{51}}%
\newcommand{\xmark}{\ding{55}}%

\usepackage{hhline}
\usepackage{booktabs}

\definecolor{applegreen}{rgb}{0.55, 0.71, 0.0}
\definecolor{cadmiumred}{rgb}{0.89, 0.0, 0.13}
\definecolor{azure}{rgb}{0.0, 0.5, 1.0}
\definecolor{amber}{rgb}{1.0, 0.75, 0.0}
\definecolor{darkorchid}{rgb}{0.6, 0.2, 0.8}
\newcommand{\changesecond}[1]{{\leavevmode\color{black}#1}}
\newcommand{\change}[1]{{\leavevmode\color{black}#1}}

\newcommand\blfootnote[1]{%
	\begingroup
	\renewcommand\thefootnote{}\footnote{#1}%
	\addtocounter{footnote}{-1}%
	\endgroup
}

\title[Chart Question Answering: State of the Art and Future Directions]%
{Chart Question Answering: State of the Art and Future Directions}

\author[E. Hoque \& P. Kavehzadeh \& A. Masry]
{\parbox{\textwidth}{\centering E. Hoque$^{1}$
		and P. Kavehzadeh$^{1}$ and A. Masry$^{1}$
	}
	\\
	{\parbox{\textwidth}{\centering $^1$Intelligent Visualization Lab, York University, Toronto, Canada
		}
	}
}

\begin{document}
	
	
	\maketitle
	
	\begin{abstract}
		
		Information visualizations such as bar charts and line charts are very common for analyzing data and discovering critical insights. Often people analyze charts to answer questions that they have in mind. Answering such questions can be challenging
		as they often require a significant amount of perceptual and cognitive effort. Chart Question Answering (CQA) systems typically 
		take a chart and a natural language question as input and automatically generate the answer to facilitate visual
		data analysis. Over the last few years, there has been a growing body of literature on the task of CQA. In this survey, we systematically review the current state-of-the-art research focusing on the problem of chart question answering. We provide a taxonomy by identifying several important dimensions of the problem domain including possible inputs and outputs of the task and discuss the advantages and limitations of proposed solutions. We then summarize various evaluation techniques used in the surveyed papers. Finally, we outline the open challenges and future research opportunities related to chart question answering.

		\ccsdesc[500]{Human-centered computing~Visualization}
		\ccsdesc[500]{Computing methodologies~Natural language processing}
		\printccsdesc
		
	\end{abstract}
	
	\blfootnote{This is the accepted version of the following article: Enamul Hoque, Parsa Kavehzadeh and Ahmed Masry, Chart question answering: State of the art and future directions, Journal of Computer Graphics Forum (Proc. Eurovis) , 2022, which has been published in final form at
		http://onlinelibrary.wiley.com. This article may be used for non-commercial purposes in accordance with
		the Wiley Self-Archiving Policy [http://olabout.wiley.com/WileyCDA/Section/id-820227.html].}
	
	\vspace{-10mm}

	\section{Introduction}
	
\begin{figure}[h]
		\includegraphics[width=\linewidth]{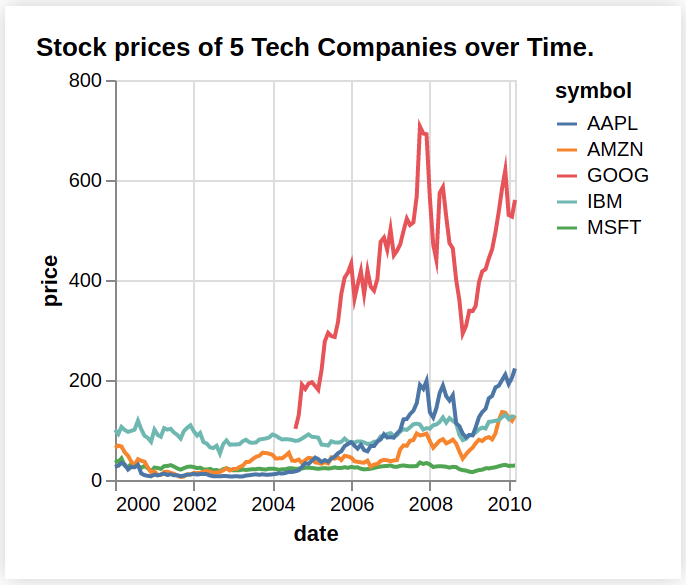}
		Question: \textit{\textbf{When the difference between the Apple and Google stocks was the highest?}} \\
		Answer: \textit{\textbf{2008}}
		\caption{ An example of a natural language question about a line chart that shows stock prices of some tech companies over time.
			\\} 
		\label{fig:stockprice}
		\vspace{-4mm}
	\end{figure}

	Information visualizations such as bar charts and line charts are commonly used for analyzing data and making informed decisions. To analyze data, often people ask various complex questions about charts~\cite{kim2020}.
	However, answering such questions about charts is not always easy. In order to answer complex questions, one must 
	have various analytical skills and would need to  
	combine various low-level operations (e.g. retrieve values from bars, find extremes, aggregate values) which can be mentally taxing.  For example, to answer the question \textit{``When the difference between the Apple and Google stocks was the highest?''} in Figure \ref{fig:stockprice}, one needs to compute the differences between all data points of two red and blue lines and then find the highest one.
	
	
	\begin{figure*}[h]
		\includegraphics[width=0.95\textwidth]{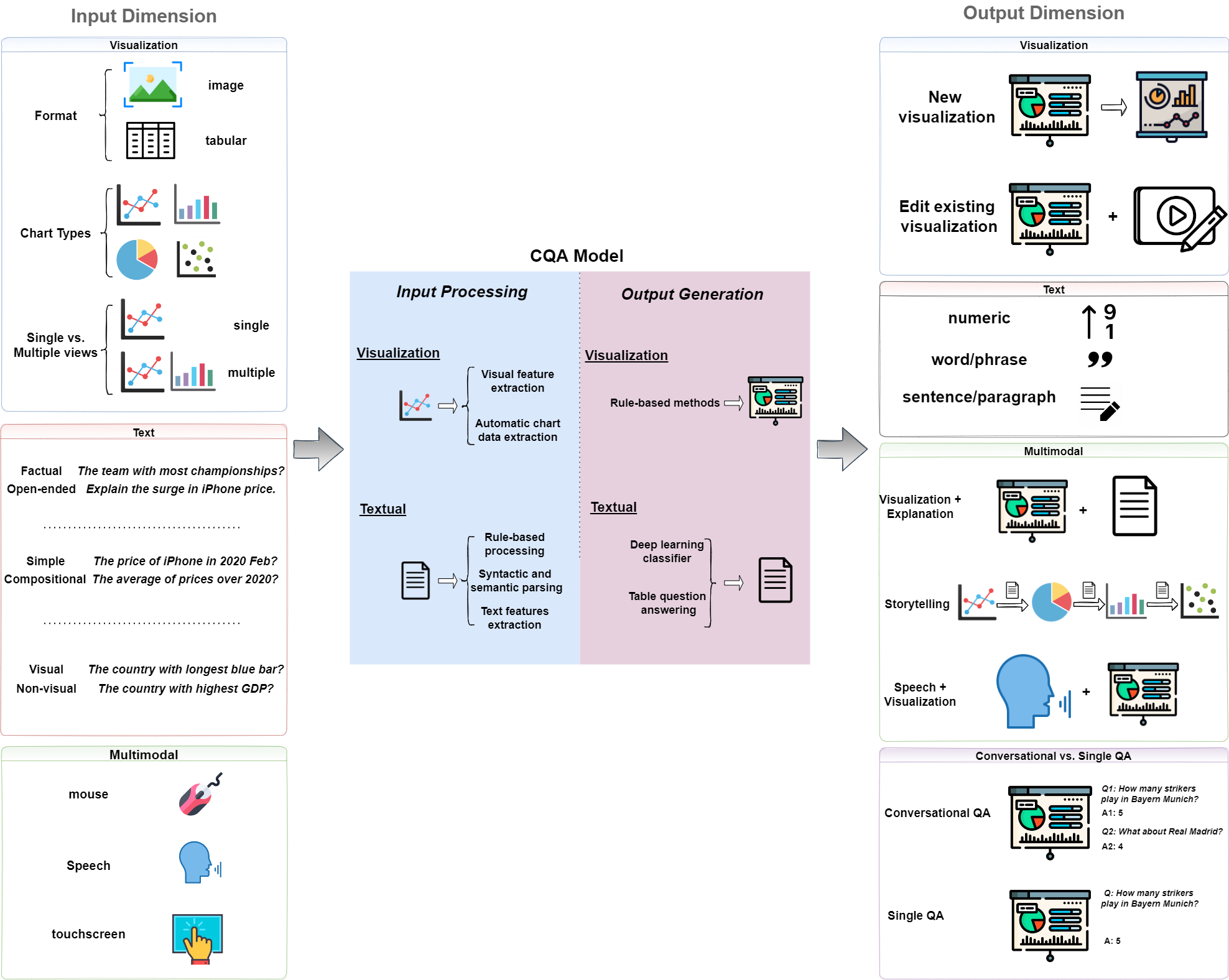}
		\centering
		\caption{An overview of the problem space of chart question answering, covering key categories of input and output dimensions}.
		A particular CQA problem setup may involve one or \change{more} categories of \textit{input} and one or more categories of \textit{output} dimensions. 
		
		\label{fig:introduction}
	\end{figure*}
	
	The goal of a chart question answering system is to automatically answer a natural language question about a chart to facilitate visual data analysis. The CQA problem belongs to the area of natural language
	interfaces (NLI) for visualizations, which has recently received a lot of attention in the research community~\cite{evizeon, orko} as well as in the industry \cite {ask-data, ferrari2016introducing}. Unlike traditional approaches which typically support visual data analysis using mouse-based interactions, NLIs allow people to express their complex information needs easily through text or speech, thus lowering the threshold of analytical skills required for analyzing data.  Moreover, it can significantly reduce the time and mental efforts\change{. It is well known that performing elementary perceptual tasks like judging the length or area of a chart is not always easy~\cite{cleveland1985graphical}. Similarly, performing various arithmetic and logical operations such as comparisons and aggregations involve cognitive activities. Automatic question answering can perform these operations on behalf of users to reduce cognitive overload. }
	Finally, CQA can enhance chart accessibility, where blind people can comprehend charts by asking questions. \change{ For example, blind people can get an overview of a chart using caption and data sonification and then they can ask simple questions~\cite{sharif2022voxlens} to further understand the given chart.}
	
	
	Automatic chart question answering is a challenging task because of the richness and ambiguities of natural language and complex reasoning that may be required to predict the answer~\cite{eviza}. The problem is also very inter-disciplinary in nature, falling in the intersection of  information visualization, natural language processing, and human computer interactions. Some early approaches on NLIs and chart question answering
	applied natural language processing  techniques by largely depending on heuristics or grammar-based parsing techniques ~\cite{eviza, orko, evizeon, datatone}. 
	Recently, there have been also some attempts for applying deep learning models for understanding natural language queries about visualizations~\cite{leafnet, stlcqa, figurenet}. Several benchmark datasets have also been developed to evaluate the effectiveness of these approaches \cite{dvqa, figureqa, 
		leafnet}.

	While there is a growing body of literature in the space of NLIs for visualizations in general and chart question answering in particular, there has not been any comprehensive survey on the topic of CQA. Some initial efforts summarize the prior works and highlight future challenges~\cite{srinivasan2020ask, vNLI-survey}, however, they cover NLIs for visualizations broadly without focusing on the specific chart question answering task.  There also exist survey papers on question answering in other domains such as question answering on text data~\cite{zhu2021retrieving}, image with scenes and objects~\cite{huang2021survey, zhang2019information} and knowledge base~\cite{lan2021complex}. However, to our knowledge there is no survey paper specifically focusing on chart question answering. This survey aims to fill that gap by systematically reviewing the state of the art on CQA and introducing a taxonomy of the problem.

	In this paper, we present a formal taxonomy of the chart question answering task and categorize the existing solutions and evaluation techniques as outlined in Section~\ref{sec:outline}. To narrow down the scope of the survey, we only focus on research works that take a question about a chart as input and produce the answer as output. We consider chart question answering as a sub-problem of developing NLIs (\cite{eviza, orko}), where NLIs focus on a broader range of tasks (e.g., manipulating visualizations through natural language commands, web-search like short queries) in addition to question answering~\cite{srinivasan2021collecting}.
	Chart question answering is also related to natural language generation (NLG) for visualizations~\cite{charttotext} and narrative storytelling as a CQA system may answer a question by generating visualizations and texts~\cite{storytelling-survey}. In order to describe the problem and the design space of proposed solutions, we identify several important dimensions of the problem domain including possible inputs and outputs of the task (see Figure~\ref{fig:introduction}). We also categorize the existing solutions and grouped the evaluation techniques used in the surveyed papers. Finally, we outline the open challenges and future research opportunities in this domain.

	\section{Structure of the Paper and Outline}
	\label{sec:outline}
	
	To comprehensively survey the CQA domain and develop a taxonomy, we carried out exhaustive searches through Google Scholar using key phrases such as ``chart question answering'', ``figure question answering'', ``natural language interfaces for visualizations'', and ``question answering with data visualizations''. In addition, we browse through publications from the venues related to information Visualization and natural language processing fields in the last five years. We then reviewed the relevant papers and applied an iterative coding approach to discover the main categories which helped us to  characterize the problem space. In particular, we analyze the CQA problem across the Input and output dimensions and review existing Evaluation techniques. Figure \ref{fig:introduction} visually summarizes the design space of the CQA problem along with major categories that we identified through our survey. Below we summarize the structure of the survey inspired by this categorization scheme. 
	
	

	\begin{figure*}
		\begin{subfigure}[t]{.28\textwidth}
			\centering
			\includegraphics[width=\linewidth]{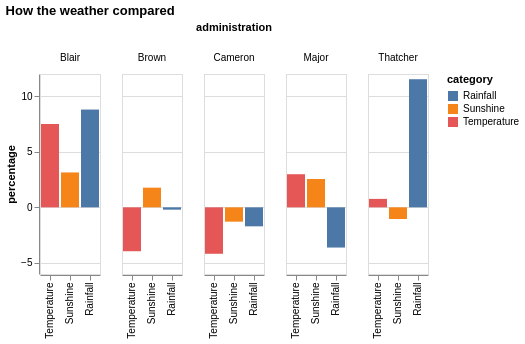}
			\caption{Vertical Bar Chart}
			\label{sfig:weather}
		\end{subfigure}
		\begin{subfigure}[t]{.4\textwidth}
			\centering
			\includegraphics[width=\linewidth]{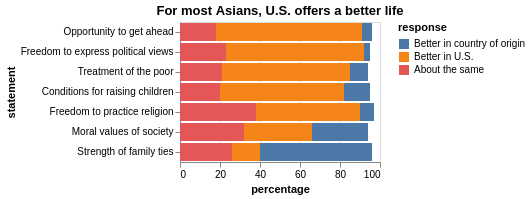}
			\caption{Horizontal Bar Chart}
			\label{sfig:life}
		\end{subfigure}
		\begin{subfigure}[t]{.23\textwidth}
			\centering
			\includegraphics[width=\linewidth]{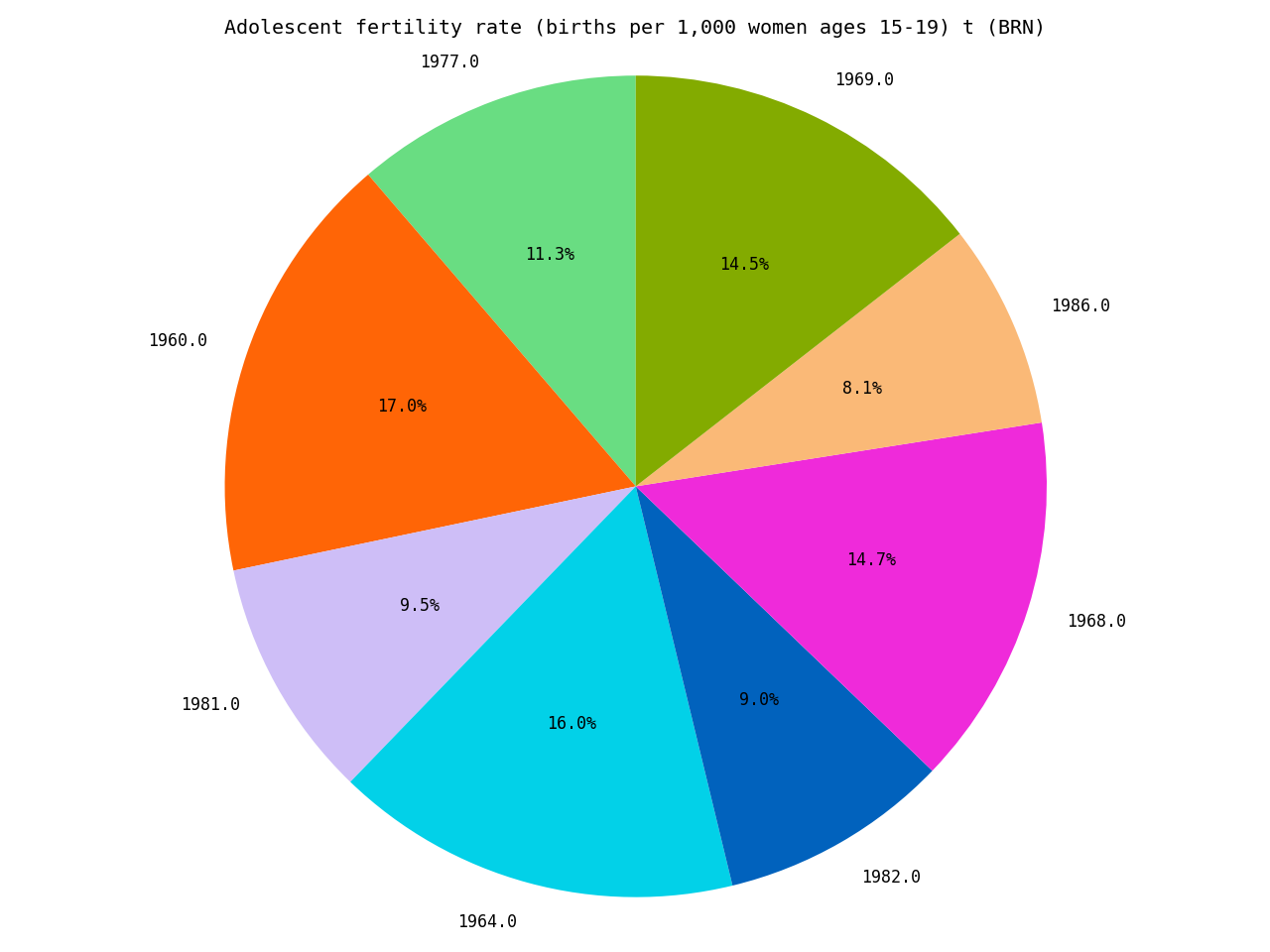}
			\caption{Pie Chart}
			\label{sfig:community}
		\end{subfigure}
		\begin{subfigure}[t]{.33\textwidth}
			\centering
			\includegraphics[width=\linewidth]{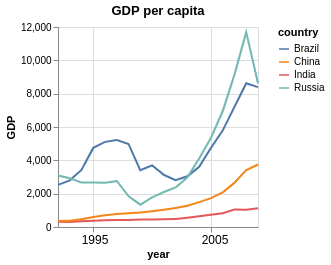}
			\caption{Line Chart}
			\label{sfig:gdp}
		\end{subfigure}
		\begin{subfigure}[t]{.33\textwidth}
			\centering
			\includegraphics[width=\linewidth]{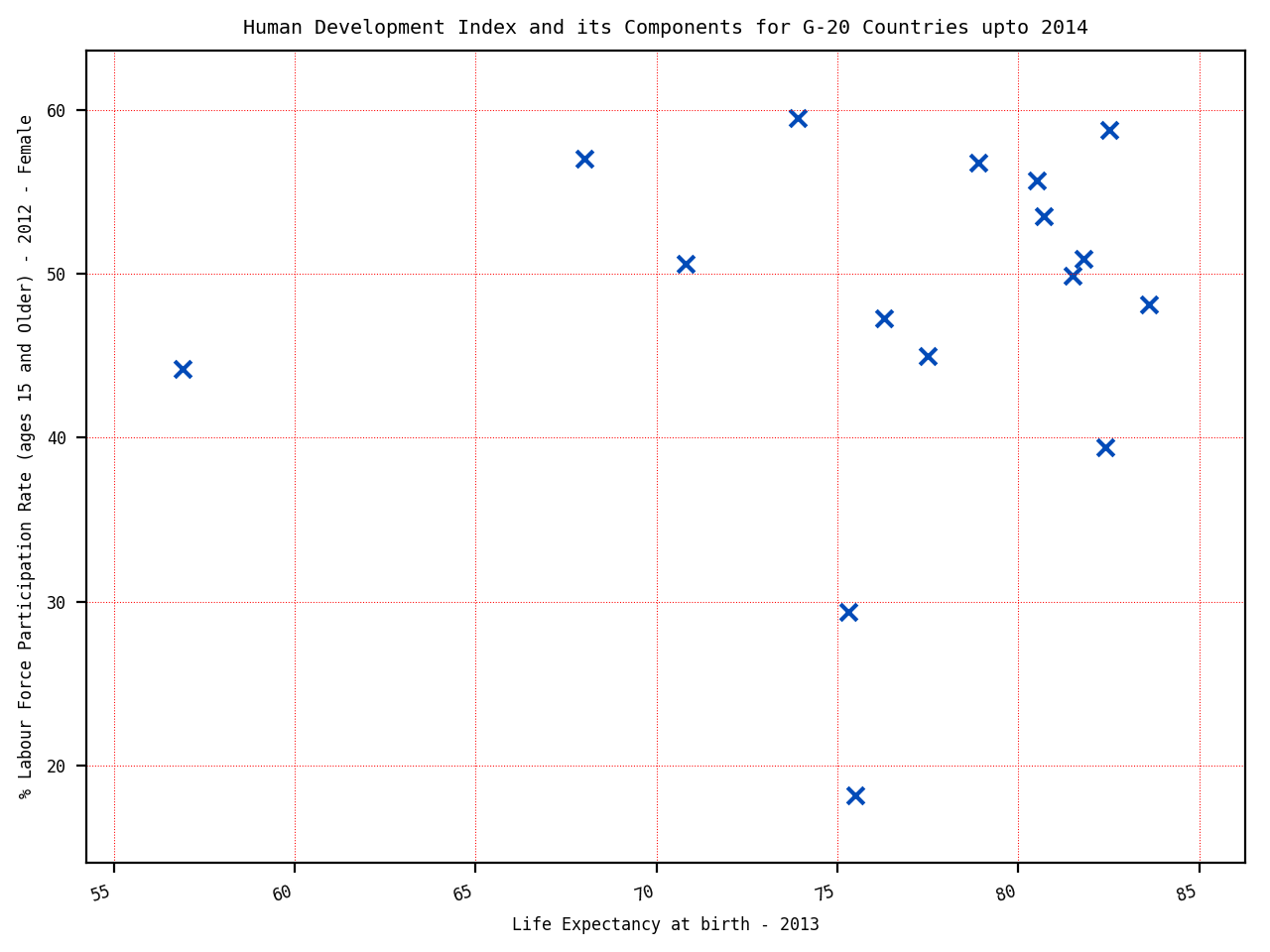}
			\caption{Scatter Plot}
			\label{sfig:dinner}
		\end{subfigure}%
		\begin{subfigure}[t]{.33\textwidth}
			\centering
			\includegraphics[width=\linewidth]{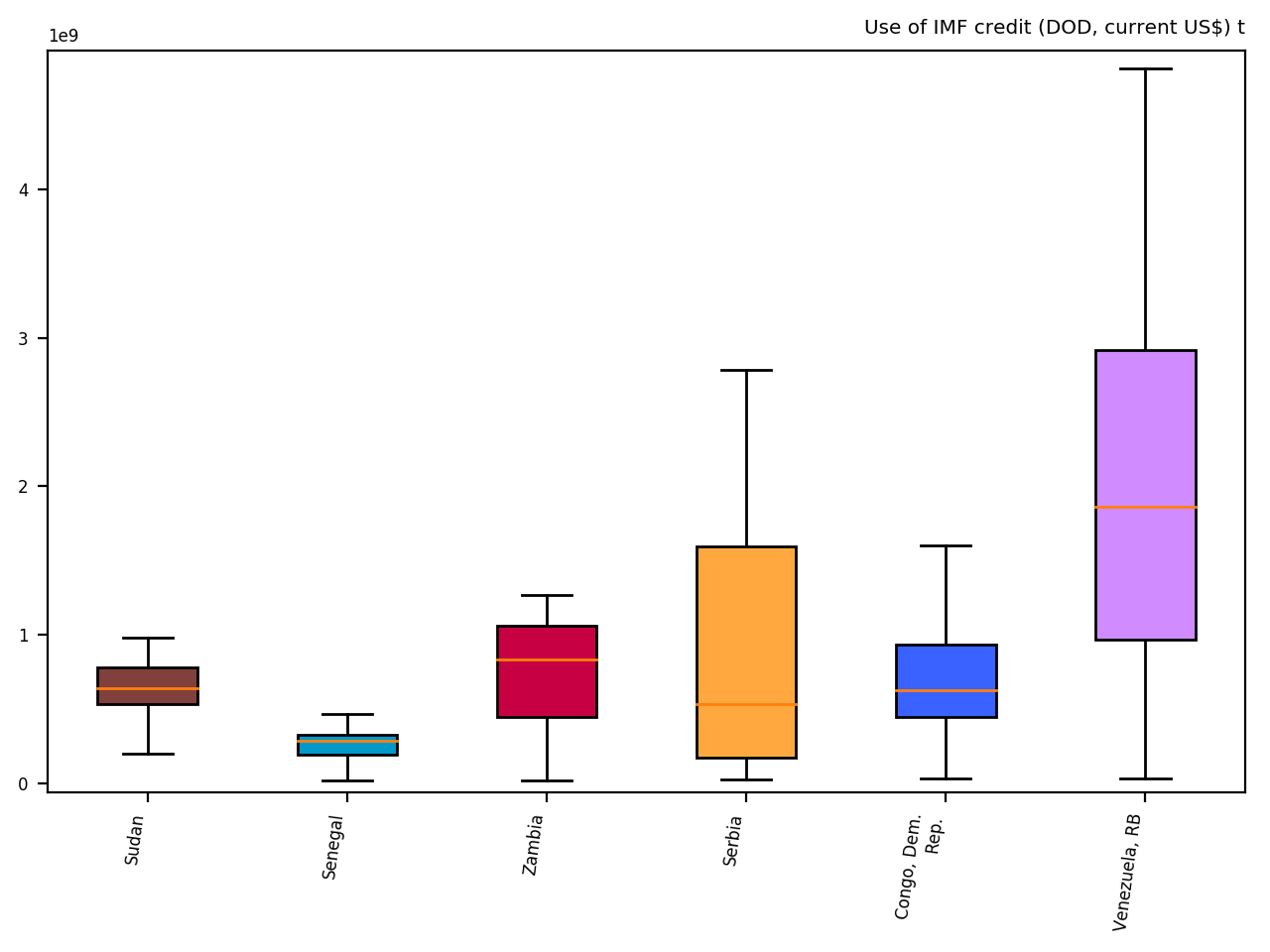}
			\caption{Box Chart}
			\label{sfig:agerecovery}
		\end{subfigure}
		\caption{Examples of different types of input charts (plots a, b, d are adapted from \cite{kim2020} and plots c, e, f are adapted from \cite{leafnet}) 
		}
		\label{fig:chartplots}
	\end{figure*}
	
	\begin{table*}[t]
		\centering
		\caption{Different types of question inputs and possible answers. Each question is referring to a particular chart in Figure \ref{fig:chartplots}, as specified at the end of the question.
		}
		\scalebox{0.68}{
			\begin{tabular}{|l|c c|c c c|}
				\toprule
				\multirow{2}{*} & \multicolumn{5}{c|}{\textbf{Answer Types}} \\
				\cmidrule{2-6}
				
				{\textbf{\makecell{Question \\ Types}}} & \multicolumn{2}{c|}{\textbf{Fixed Vocabulary}} & \multicolumn{3}{c|}{\textbf{Open Vocabulary}} \\
				\cmidrule{2-6}
				& \textbf{Numeric} & \textbf{Word(s)/Phrase(s)} & \textbf{Numeric} & \textbf{Word(s)/Phrase(s)} & \textbf{\makecell[l]{Explanatory \\ Sentence(s)}}\\
				\cmidrule{1-6}
				\textbf{Factual} & \multicolumn{1}{l}{\makecell[|l]{How many years have fertility\\rate more than 11 percent? \ref{sfig:community}}}
				& \multicolumn{1}{l|}{\makecell[|l]{Which years have more \\ than 11 percent of fertility? \ref{sfig:community}}} & \multicolumn{1}{l}{\makecell[l]{How much is life expectancy's\\mean bigger than labour force's? \ref{sfig:dinner}}}
				& \multicolumn{1}{l}{\makecell[l]{ 
						To which continent Venezuela\\ belongs to?
						\ref{sfig:agerecovery}}}
				& -\makecell{}
				
				\\
				\cmidrule{2-6}
				\textbf{Open-ended}  & - & - & -& - & \makecell[l]{Explain the surge of Russia's GDP \\ in the end of 1990's. \ref{sfig:gdp}} \\
				\cmidrule{1-6}
				
				\textbf{Visual} & \multicolumn{1}{l}{\makecell[l]{What is the median \\of smallest box? \ref{sfig:agerecovery}}} & \multicolumn{1}{l}{\makecell[l]{Which administration has the\\longest blue component?\ref{sfig:weather}}} & \multicolumn{1}{|l}{\makecell[l]{What is the ratio of the longest \\ orange bar to the shortest? \ref{sfig:weather}}} & \multicolumn{1}{l}{\makecell[l]{
						To which continent the red \\colored box belong to?
						\ref{sfig:agerecovery}}} & \multicolumn{1}{l|}{\makecell[l]{Why Thatcher has the longest \\ blue component? \ref{sfig:weather}}} \\
				\cmidrule{2-6}
				
				\textbf{Non-visual} & \multicolumn{1}{l}{\makecell[l]{What is the median of \\ Sudan? \ref{sfig:agerecovery}}} & \multicolumn{1}{l|}{\makecell[l]{Which statement satisfies Asians \\ most regarding living in U.S.? \ref{sfig:life}}} & \multicolumn{1}{l}{\makecell[l]{What is the ratio of Sudan's \\ median to Serbia's? \ref{sfig:agerecovery}}} & \multicolumn{1}{l}{\makecell[l]{What is the population of the \\ country with highest gdp? \ref{sfig:gdp}}} & \multicolumn{1}{l|}{\makecell[l]{Why higher life expectancy \\ included higher female labour? \ref{sfig:dinner}}}\\
				\cmidrule{1-6}
				
				\textbf{Simple} & \multicolumn{1}{l}{\makecell[l]{What is the percentage of \\ fertility in 1982? \ref{sfig:community}}} & \multicolumn{1}{l|}{\makecell[l]{Which country experienced the \\ gdp higher than 10,000? \ref{sfig:gdp}}} & \multicolumn{1}{l}{\makecell[l]{What is the percentage of \\ rainfall in Blair? \ref{sfig:weather}}} & \multicolumn{1}{l}{\makecell[l]{
						Was the blue line increasing or \\decreasing from 2000 to 2005?
						\ref{sfig:gdp}}} & \multicolumn{1}{l|}{\makecell[l]{What is the definition \\ of GDP? \ref{sfig:gdp}}}\\
				\cmidrule{2-6}
				
				\textbf{Compositional} & \multicolumn{1}{l}{\makecell[l]{What is the overall percentage of \\ fertility in 1982 and 1968? \ref{sfig:community}}} & \multicolumn{1}{l|}{\makecell[l]{Which country has the highest \\ median? \ref{sfig:agerecovery}}} & \multicolumn{1}{l}{\makecell[l]{What is the mean of total \\ life expectancy rates? \ref{sfig:dinner}}} & \multicolumn{1}{l}{\makecell[l]{In which continent is the country\\ with the highest GDP last year? \ref{sfig:gdp}}} & \multicolumn{1}{l|}{\makecell[l]{Why is China's GDP \\ less than Brazil's? \ref{sfig:gdp}}}\\
				\bottomrule
				
		\end{tabular}}
		\label{tab:examples}
	\end{table*}

	\begin{table*}[]
		\setlength\extrarowheight{2pt}
		\centering
		\caption{
			The table summarizes how existing research papers on CQA cover different categories of input and output dimensions. In the left most column, the papers coded with \textcolor{azure}{blue} color  represent those works that solely focus on CQA, while the ones with \textcolor{magenta}{magenta} color 
			are the works that do cover some question answering aspects but focus more broadly on supporting natural language commands and web-search like queries for interactions with visualizations (such as filtering, sorting, zooming, and highlighting).
		}
		\label{my-label}
		
		\scalebox{0.73}{\begin{tabular}{|l|l|l|l|l|l|l|l|l|l|l|}
				
				\cline{1-11}
				
				\multirow{4}{*}{} & \multicolumn{10}{c|}{Dimensions}   \\ \cline{2-11} 
				
				& \multicolumn{5}{c|}{Input} & \multicolumn{5}{c|}{Output} \\ \cline{2-11}
				
				& \multicolumn{3}{c|}{Text} & Visualization & \multirow{3}{*}{Multimodal} & \multicolumn{2}{c|}{Text} & Visualization & \multirow{3}{*}{Multimodal} & \change{\multirow{3}{*}{Conversational}}\\ 
				\cline{2-5}
				\cline{7-9}

				Related Papers & \multirow{2}{*}{\makecell{Factual/\\Open-ended}} & \multirow{2}{*}{\makecell{Visual/\\Non-visual}} & \multirow{2}{*}{\makecell{Simple/\\Compositional}} & \multirow{2}{*}{\makecell{Bitmap/\\Data Table}} &  & \makecell{Fixed\\Vocab} & \makecell{Open\\Vocab} & \multirow{2}{*}{\makecell{New/\\Existing}} & &  \\ 
				
				\cline{7-8} 
				
				& & & & &  & \makecell{Numeric/\\Word} & \makecell{Numeric/\\Word/\\Sentence} & & & \\
				
				\hline
				
				\rowcolor{gray!30}
				\textcolor{azure}{LeafNet}\cite{leafnet} & \cmark/\xmark & \cmark/\cmark & \cmark/\cmark & \cmark/\xmark & \xmark & \cmark/\cmark & \xmark/\xmark/\xmark & \xmark/\xmark & \xmark & \xmark\\ \cline{2-11}
				
				\textcolor{azure}{STL-CQA}\cite{stlcqa} & \cmark/\xmark & \cmark/\cmark & \cmark/\cmark & \cmark/\xmark & \xmark & \cmark/\cmark & \xmark/\xmark/\xmark & \xmark/\xmark & \xmark & \xmark \\ \cline{2-11} 
				
				\rowcolor{gray!30} 
				\textcolor{azure}{DVQA}\cite{dvqa} & \cmark/\xmark & \cmark/\cmark & \cmark/\cmark & \cmark/\cmark & \xmark &    \cmark/\cmark & \xmark/\xmark/\xmark & \xmark/\xmark & \xmark & \xmark \\ \cline{2-11}

				\textcolor{azure}{PReFIL}\cite{prefil} & \cmark/\xmark & \cmark/\cmark & \cmark/\cmark & \cmark/\xmark & \xmark   & \cmark/\cmark & \xmark/\xmark/\xmark & \xmark/\xmark  & \xmark & \xmark \\ \cline{2-11}
				
				\rowcolor{gray!30} 
				\textcolor{azure}{Kim et al.}\cite{kim2020} & \cmark/\xmark & \cmark/\cmark & \cmark/\cmark  & \xmark/\cmark & \xmark & \cmark/\cmark & \cmark/\cmark/\cmark & \xmark/\xmark & \xmark& \xmark \\ \cline{2-11} 
				
				\textcolor{azure}{FigureNet}\cite{figurenet} & \cmark/\xmark & \cmark/\cmark & \cmark/\cmark & \cmark/\cmark & \xmark   & \xmark/\cmark & \xmark/\xmark/\xmark & \xmark/\xmark  & \xmark & \xmark \\ \cline{2-11}
				
				\rowcolor{gray!30}
				\textcolor{azure}{Affinity}\cite{affinity} & \cmark/\xmark & \cmark/\cmark & \cmark/\cmark & \cmark/\xmark & \xmark   & \cmark/\cmark & \xmark/\xmark/\xmark & \xmark/\xmark  & \xmark & \xmark \\ \cline{2-11}
				
				\textcolor{azure}{FigureQA}\cite{figureqa} & \cmark/\xmark & \cmark/\cmark & \cmark/\cmark & \cmark/\cmark & \xmark   & \cmark/\cmark & \xmark/\xmark/\xmark & \xmark/\xmark  & \xmark & \xmark \\ \cline{2-11}
				
				\rowcolor{gray!30}

				\textcolor{azure}{PlotQA}\cite{plotqa} & \cmark/\xmark & \cmark/\cmark & \cmark/\cmark  & \cmark/\xmark & \xmark  & \cmark/\cmark & \cmark/\cmark/\xmark & \xmark/\xmark  & \xmark & \xmark \\ \cline{2-11}
				
				\textcolor{azure}
				{ChartQA}\cite{Masry2022ChartQAAB}
				& \cmark/\xmark & \cmark/\cmark & \cmark/\cmark & \cmark/\cmark & \xmark & \cmark/\cmark & \cmark/\cmark/\cmark  & \xmark/\xmark  & \xmark & \xmark \\ \cline{2-11}  
				
				\specialrule{.8pt}{0pt}{0pt}%
				\cline{1-2}
				\cline{2-11}
				\rowcolor{gray!30}
				{\color{magenta} Eviza}\cite{eviza} & \cmark/\xmark & \xmark/\cmark & \cmark/\cmark  & \xmark/\cmark & \cmark & \xmark/\xmark & \xmark/\xmark/\xmark & \xmark/\cmark  & \xmark & \cmark \\ \cline{2-11}

				{\color{magenta} Evizeon}\cite{evizeon} & \cmark/\xmark & \cmark/\cmark & \cmark/\cmark  & \xmark/\cmark & \cmark & \xmark/\xmark &  \xmark/\xmark/\xmark & \xmark/\cmark  & \xmark & \cmark \\\cline{2-11}
				
				\rowcolor{gray!30}
				{\color{magenta} DataTone}\cite{datatone} & \cmark/\xmark & \xmark/\cmark & \cmark/\cmark  & \xmark/\cmark & \cmark & \xmark/\xmark & \xmark/\xmark/\xmark & \xmark/\cmark  & \xmark & \xmark \\ \cline{2-11}

				{\color{magenta} Orko}\cite{orko} & \cmark/\xmark & \xmark/\cmark & \cmark/\cmark  & \xmark/\cmark & \cmark & {\xmark/\xmark} & \xmark/\xmark/\xmark & \xmark/\cmark  & \cmark & \cmark \\ \cline{2-11}
				
				\rowcolor{gray!30}
				{\color{magenta} FlowSense}\cite{flowsense} & \cmark/\xmark & \xmark/\cmark & \cmark/\cmark  & \xmark/\cmark & \xmark & \xmark/\xmark & \xmark/\xmark/\xmark & \cmark/\cmark  & \xmark & \cmark \\ \cline{2-11}

				{\color{magenta} NL4DV}\cite{nl4dv} & \cmark/\xmark & \xmark/\cmark & \cmark/\cmark & \xmark/\cmark & \xmark & \xmark/\xmark & \xmark/\xmark/\xmark & \cmark/\xmark  & \xmark & \xmark \\ \cline{2-11}
				
				\rowcolor{gray!30}
				\textcolor{magenta}{ADVISor}\cite{advisor} & \cmark/\xmark & \xmark/\cmark & \xmark/\cmark & \xmark/\cmark & \xmark & \xmark/\xmark & \xmark/\xmark/\xmark & \cmark/\xmark  & \xmark & \xmark \\ \cline{2-11}

				\textcolor{magenta}{NL2VIS} \cite{luo2021natural} & \cmark/\xmark & \xmark/\cmark & \cmark/\cmark & \xmark/\cmark & \xmark & \xmark/\xmark & \xmark/\xmark/\xmark & \cmark/\xmark  & \xmark & \xmark \\ \cline{2-11}
				\hline
				
		\end{tabular}}
		\label{tab:papers}
	\end{table*}
	
	\noindent \textbf{(1) Inputs:} Chart question answering systems may take a variety of visualization, text, and multimodal inputs as illustrated in Figure~\ref{fig:introduction} (left). Most papers we reviewed on the CQA task considered that the system takes a visualization and a natural language question about it as input~\cite{leafnet,stlcqa, dvqa, prefil, kim2020, figurenet, plotqa}. There are also some works on natural language interfaces that enable \textit{multimodal} inputs by combining touch, speech and other modalities (e.g. Orko~\cite{orko, srinivasan2020inchorus}). The given question can be categorized in various ways, for example, based on complexity (e.g. simple vs. complex), or whether it refers to visual attributes of graphical marks in a chart (e.g. visual vs. non-visual). Similarly, charts can be presented in different formats (e.g. bitmap image vs. a SVG chart with access to underlying data table) as well as with various types (e.g. bar charts, line charts). \change{When the underlying data table is available, the input chart can be presented in a declarative specifications such as using Vega-Lite \cite{vegalite}. Another possible problem variation could take multiple views as input with several underlying data tables, \changesecond{ possibly created by visualization recommendation systems\cite{hu2019vizml}}.}
	In Section \ref{sec:input}, we will dive deep into the methods used for processing each input type in a CQA system.
	
	\noindent \textbf{(2) Outputs:} Like inputs, outputs of a CQA system can be presented in different forms as shown in Figure~\ref{fig:introduction} (right). Most CQA systems output textual answers to the given query but they can be characterized into different types depending on whether the answer comes from a fixed vocabulary with limited possible answers like `yes' and `no' (e.g., ~\cite{figureqa, dvqa, leafnet})
	or from an open vocabulary with various possible answers (e.g., ~\cite{plotqa, kim2020}). Some other natural language interfaces produce a visualization as output~\cite{datatone} or highlight answers in an existing visualizations~\cite{eviza} or even combine both visualization and text/audio as multimodal output~\cite{orko}. \change{Some natural language interfaces also produce a response to the current question in the context of the  previous questions  asked by the user. Moving beyond the single query-response paradigm, these systems show the advantage of improving the flow of analytical conversation.
		We will discuss different possible types of output in CQA in Section \ref{sec:output}.}
	
	
	\noindent \textbf{\change{(3)} Evaluation: } \change{In recent years, researchers have released several benchmark datasets to evaluate the performance of the question answering ~\cite{figureqa, dvqa, plotqa} and NLI \cite{sql2vis2, text2sql} systems.} Other works have focused on conducting user studies to measure the performance and subjective feedback from participants to understand the effectiveness and limitations of different prototypes (e.g., ~\cite{eviza, orko}). In section \ref{sec:evaluation}, we will critically review these benchmark datasets and user study methods \changesecond{with respect to input and output dimensions} to highlight the challenges and future directions in evaluating chart question answering systems.

	\begin{figure*}[h]
		\begin{subfigure}[t]{0.51\textwidth}
			\centering
			\includegraphics[width=\linewidth]{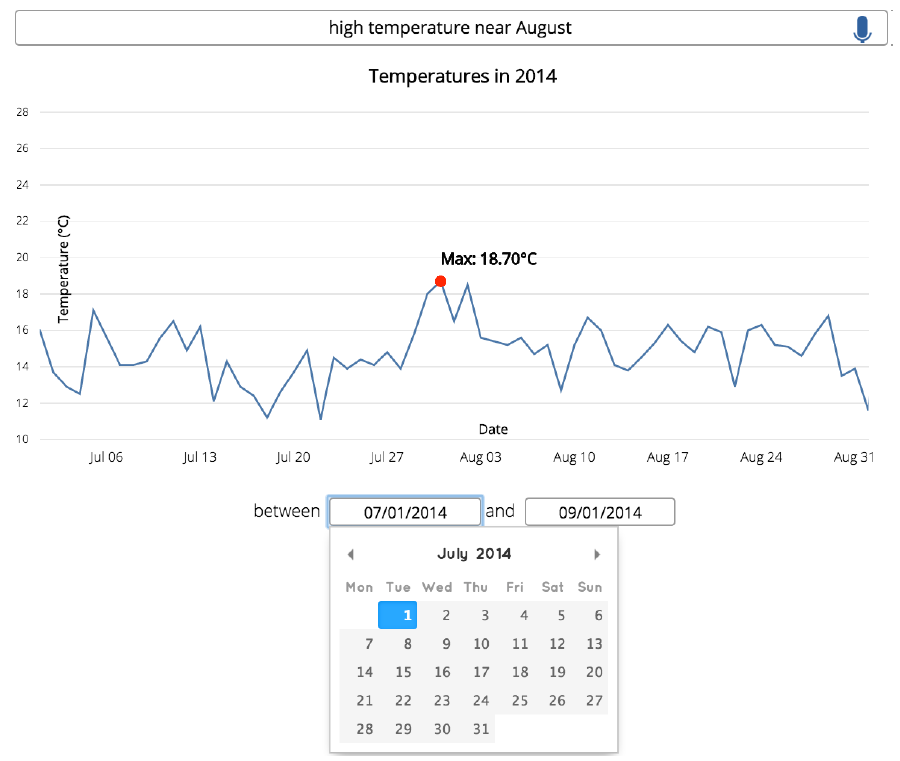}
			\caption{}
			\label{subfig:outputvisa}
		\end{subfigure}
		\begin{subfigure}[t]{0.46\textwidth}
			\centering
			\includegraphics[width=\linewidth]{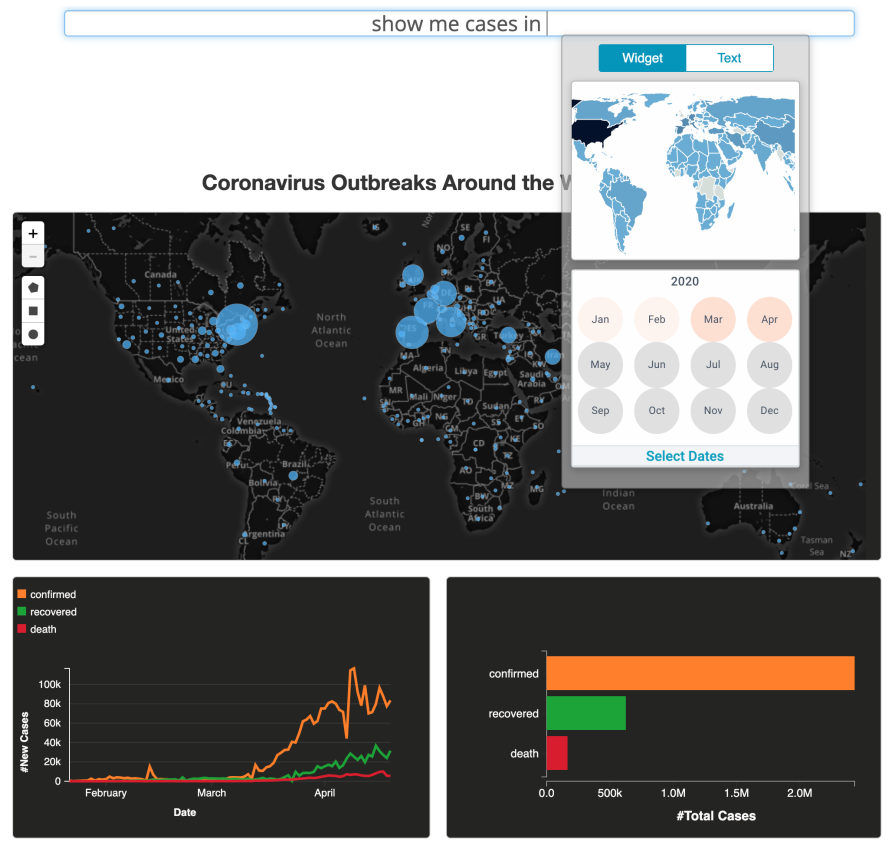}
			\caption{}
			\label{subfig:outputvisb}
		\end{subfigure}
		\caption{
			Examples of how text and interactive visualization can be combined for inputting a query. In Figure \ref{subfig:outputvisa}, the user types a vague query and then resolves the ambiguity by  selecting more specific date from a calendar widget\cite{eviza}. In  Figure \ref{subfig:outputvisb}, as the user types a partial query, two visualizations are popped up with  previews of the dataset, thus supporting the user in completing the query\cite{sneak-peak}. 
		}
		\label{fig:outputvis}
	\end{figure*}
	

	\section{Input Dimension}
	\label{sec:input}
	
	In this section, we discuss the possible input dimensions of a CQA system including \textit{Text}, \textit{Visualization}, and \textit{multimudal} inputs and review how existing research works analyze these inputs to answer the question.
	
	\subsection{Textual}
	In chart question answering, people express their information needs through a variety of textual queries as shown in Table~\ref{tab:examples}. Note that these categories are not orthogonal; for example, a question can be factual, visual, and compositional simultaneously. We discuss the different types of textual queries along with how current CQA systems handle them below.
	
	\noindent{\textbf{ $\bullet$ Factual vs. Open-ended} Most existing works focus on answering questions that require \textbf{factual} answers (e.g. \textit{``Which country has the highest GDP?''}) \cite{leafnet, kim2020, figurenet, dvqa}. Factual questions require the system to compute the answer by analyzing the question and the chart and then subsequently performing some arithmetic and logical operations (e.g., compare values, find extremes). Several works have attempted to understand and analyze such questions using a combination of heuristic approaches and syntactic parsing techniques~\cite{eviza, orko, nl4dv}. Others have focused on leveraging various machine learning approaches for table question answering~\cite{kim2020, ahmed-workshop-2021} and feature extraction from chart images~\cite{leafnet, figurenet, dvqa}. For example, several research works utilize the recurrent neural model (RNN) with Long short-term memory (LSTM) architecture to encode the question\cite{leafnet, figurenet, plotqa, prefil, affinity, dvqa}. However, more recently transformer architecture has been found to be more effective than RNN for long input sequences, which inspired some researchers to adopt such architecture for answering factual questions with charts~\cite{stlcqa, ahmed-workshop-2021, Masry2022ChartQAAB, advisor}. Unlike factual questions, \textbf{open-ended} questions are exploratory in nature (e.g., \textit{``Why the country with the highest GDP changed in 2016?''}) and typically the expected answer is an explanatory text. Generating such explanatory answers is very challenging and to our knowledge, there is no existing work that explores this problem. The closest works to this problem are the ones that automatically summarize the key insights from charts as text \cite{demir-etal-2012-summarizing, Andrea-carenini, charttotext, chart-to-text-acl}. For example, Obeid and Hoque \cite{charttotext} adopted a transformer-based model to generate an explanatory text describing the chart. Future works may explore such data-to-text generation approaches by taking the question as additional input.
		
		\noindent {\textbf{$\bullet$ Visual vs. Non-visual}
			Visual questions include reference to visual attributes such as \textit{color}, \textit{height}, and \textit{length} of graphical marks (e.g., \textit{bars}) in the chart. In contrast, non-visual questions do not contain such references. For instance, the question \textit{``Which bar is the longest?''} is a visual question since it is referring to the \textit{length} attribute of the mark type \textit{bar}. Kim et al. \cite{kim2020} handled the visual question by translating it to an equivalent non-visual question through a pipeline that first recognizes all phrases in the input question that refers to marks and their visual attributes and then replaces these phrases with equivalent non-visual terms. Then they pass the question and the data table of the chart to a table parsing model to obtain the answer. Others have extracted visual features from the chart and then build classification models to answer the visual questions~\cite{figureqa, dvqa}.  
			
			\noindent {\textbf{$\bullet$ Simple vs. Compositional} 
				\change{This categorization is based on the complexity of the  analytical tasks\cite{amar2005low}} that the system needs to perform to answer the given question. To answer \textbf{simple} questions, the system needs to complete a single analytical task (e.g., retrieving a value). For instance, the question \textit{``What percentage of people are Real Madrid's fan?''} is a simple question. On the other hand, \textbf{compositional} questions involve multiple mathematical/logical operations like \emph{sum}, \emph{difference} and \emph{average}. Compositional questions are more challenging as the model needs to combine multiple operations together. For example, to answer this question ``How many students achieved over 90\%?'' from a bar chart that shows the grades of students, the system needs to perform a \textit{filtering} operation to find students who achieved over 90\% followed by applying a \textit{count} operation. To handle the compositional questions, some researchers have applied a compositional semantic parsing technique called Sempre \cite{sempre} that was originally trained on a table question answering dataset~\cite{kim2020, flowsense}. However, solving compositional questions still remains a challenging task as the accuracy is still very low for such questions. For example, Kim et al. \cite{kim2020} found the accuracy of their model to be only 37\% for compositional questions.
				
				
				\subsection{Visualization}
				The input visualization to the CQA system can have different types and storage format as shown in Figure~\ref{fig:introduction}. Below we discuss how current CQA methods process the input visualization.
				
				\noindent{\textbf{$\bullet$ Chart Types:} Most existing works on CQA focused on limited chart types.  Bar charts \cite{leafnet, figurenet, prefil, affinity, plotqa}, Line charts \cite{leafnet, prefil, affinity, plotqa}, Pie charts \cite{leafnet, figurenet, prefil, affinity}, Box plots \cite{leafnet}, Scatter plots \cite{leafnet, plotqa} are most common chart types used in recent related works. Figure \ref{fig:chartplots} shows examples of chart types. None of the CQA works in  Table~\ref{tab:papers} (coded with blue color) supports map chart even though they are commonly used. However, some NLI systems such as Eviza  support natural langauge queries with map charts~\cite{eviza}. Less common and unconventional chart types such as Parallel coordinates plot and radar charts are not supported in existing works.  
					
					\noindent{\textbf{$\bullet$ Single vs. multiple views:} 
						All the works on CQA listed in Table~\ref{tab:papers} focused on question answering with a single visualization while multiple coordinated views~\cite{javed2012exploring} and dashboards have rarely been considered ~\cite{evizeon, chowdhury2020designing}. When multiple views are introduced, there are several additional challenges that are introduced. For example, which view the user is referring in the question? How the system can combine reasoning over multiple charts? 
						Such questions are not deeply investigated yet.
						
						\noindent{\textbf{$\bullet$ Storage Format:}
							Storage format is another important aspect of the input chart. \change{Charts can be stored in different formats such as SVG vs. bitmap image.} Several works take charts in bitmap image format which are more challenging to handle because the model does not have access to the underlying data table. 
							\changesecond{They mainly utilized computer vision techniques to extract features from a bitmap chart image and then applied classification models ~\cite{figureqa, leafnet, plotqa, figurenet, stlcqa, prefil}. For example, FigureQA \cite{figureqa} extracts the features of the chart using a CNN. 
								The problem with such CNN-based feature extraction is that the model treats the chart like a natural scene without extracting individual graphical marks (e.g., bars) and the underlying texts (e.g, x-axis-labels) in the chart. To overcome this limitation, others attempted to parse the chart by employing object detection models such as Mask-RCNN \cite{maskrcnn} to detect visual and textual elements \cite{leafnet, stlcqa}. Subsequently, Optical Character Recognition (OCR) models are applied in order to dynamically encode the chart-specific tokens in the question in terms of the positional information of the textual elements in the chart image (e.g., `x-axis-label-1', `x-axis-label-2'). However, such dynamic encoding technique is prone to OCR errors and fails when the question refers to the chart texts using synonyms (e.g., `US' vs `United States'). 
								
								One key question here is how the model should reason over both the question and the visualization together to generate the answer? FigureQA used Relation Network (RN) \cite{relationalnetwork} to capture the relation between visual and textual features by  concatenating all pairs of object representations from the chart image provided by a CNN 
								with the question features \cite{figureqa}. However, the number of object pairs in RN can be very large, resulting in computational inefficiency and restricting the performance. Some researchers attempted to address the problem either by making relation features in RN more concise  \cite{affinity} or by dividing the problem into various sub-tasks\cite{figurenet}. 
								Others captured inter-relation between the chart and question features by applying attention mechanisms \cite{leafnet, dvqa} or through fusing the low and high level features from the chart image and the question in parallel to facilitate multi-stage reasoning process. However, these models ignore the chart structure (e.g., relationships between axis labels,  bars, and legends).
								STL-CQA \cite{stlcqa} attempted to better capture the relation between the chart element and the question using a multi-modal transformer-based model,  \changesecond{which can better exploit the structural properties by learning the intra-modality relationships (among the chart elements) and cross-modality relationships (between the question and the chart elements)\cite{lxmert}.
									
									
									A common limitation of all the above models is that they simply utilize the image features without extracting the underlying data table and visual encoding information of the given chart, which makes it difficult for their models to answer complex visual and compositional questions. 
									Moreover, since they are classification based models, they could only handle fixed-vocabulary type answers and can not apply mathematical operations on the chart data.

									In contrast to the above body of work, some systems 
									simplify the problem by assuming that the underlying data and visual encodings of charts are available \cite{eviza, orko, evizeon, advisor, nl4dv}. Kim et al.~\cite{kim2020} propose a pipeline that recovers the data table and encodings from an input chart to derive the corresponding Vega-Lite specification \cite{vegalite}. It then extracts the data and transforms into a flat relational table and passes it to a table question answering model. 
									If the data can be extracted from the chart, leveraging the pre-trained transfer-based model for table question answering such as TaPas \cite{tapas} or converting natural language question to SQL query \cite{text2sql} could be effective for CQA. 
									However, directly applying table question answering method on chart data is not enough, rather it is necessary to effectively combine the chart features with the data table. Masry et al. \cite{Masry2022ChartQAAB} takes an initial step in this direction by extracting both the visual features extracted using ViT \cite{vit} and the underlying data table of the chart encoded by TaPas \cite{tapas} and combining them using a cross modality encoder to infer the answer. Still, their model feeds the visual features and the data values separately to their model and does not relate between them. A promising direction here could be to create a better representation of the chart that combine and relate between the visual features and the data table is  needed to solve complex visual reasoning questions.
									
									}
									
									
								}

							\subsection{Multimodal}
							Multimodal interactions with visualizations have been actively explored with a  focus on using touch and pen~\cite{walny2012understanding}, body movement in front of a large display~\cite{andrews2011information}, gestures~\cite{badam2016supporting}, and coordinating between large displays and smartwatches~\cite{Horak2018}. However, none of these works considered natural language as an input modality. On the other hand, many CQA systems only considered natural language question as input without considering other modalities like touch, gesture, and body movement (e.g., \cite{figureqa, dvqa, leafnet}). 
							
							Notable exceptions are some NLIs for visualizations which combine natural language with mouse/touch as input modalities~\cite{evizeon, 
								orko, srinivasan2020inchorus}. For example, Evizeon~\cite{evizeon} and Orko\cite{orko} demonstrate multimodal interactions within a map chart and network diagram respectively. Eviza \cite{eviza} and DataTone~\cite{datatone} demonstrate how combining natural language input with mouse-based input from  ambiguity widgets can be helpful to resolve ambiguities and system's misunderstandings (see Figure \ref{subfig:outputvisa}). Sneak Pique is another system that automatically pops up autocompletion widgets from which the user can select query criteria through direct manipulation in addition the textual input~\cite{sneak-peak}. For example, in Figure \ref{subfig:outputvisb}, as the user types a textual query \textit{``show me case in''}, the system pops up a map and a calendar widgets that provide previews of data and allow users to click in the widgets to formulate the query. 
							
							Others have focused on multimodal interactions for tablet devices~\cite{kassel2018valletto, srinivasan2020inchorus}.  For example, InChorus supports multimodal interactions with visualizations on tablet devices by combining pen, touch, and speech as input modalities~\cite{srinivasan2020inchorus}. 
							Saktheeswaran et al.~\cite{saktheeswaran2020touch} ran a user study to confirm that compared to separate unimodal touch- and speech-based interfaces,  an interface with both touch and speech modalities is  preferred by participants because they felt it gave them more freedom to express the queries. Participants also felt comfortable with complementary of speech and touch modalities since whenever speech or touch was not sufficient, they could easily use another modality to illustrate what they want.  
							
							While the above body of work suggests that multimodal interaction offers strong promise, the potential challenges and opportunities in integrating natural language and other modalities have not been deeply investigated yet. In particular, it remains largely unknown how to synergistically integrate natural language interaction with other input modalities like touch, pen, and gestures in the context of different form factors ranging from large screen display, to desktop screen and mobile displays.
							\change{
								\subsection{Discussion}
								In terms of the input dimension, there are multiple possible directions for future works. First, more advanced models can be designed to address the weaknesses of current models in comprehending complex and compositional questions about charts. Simple classification approaches or QA approaches for data tables usually lack the ability to handle most real-world questions. Using the combined information from both visual features and data table 
								may overcome this issue. Second, most proposed works assumed that the underlying information of charts such as data table and visual encodings are already available; however, this assumption is not valid for most of the charts on Web that are stored in bitmap image format. Previously designed methods for extracting data values from chart images lack accuracy and devising more robust systems for chart data extraction can be helpful in improving CQA systems. 
								\change{Moreover, utilizing visual transformer models has been found to be effective for different computer vision tasks\cite{liu2021survey}, employing such neural models can be a promising direction.
								Finally, 
								existing} approaches can only handle single visualization as input, while users tend to ask complicated questions about multiple charts (e.g. dashboard) in real-world scenarios. Designing models that are capable of taking multiple views as input can be an interesting direction. We will discuss these directions in detail in Section~\ref{sec:challenges}.
							
						}

						\begin{figure}[t!]
							\includegraphics[width=.9\linewidth]{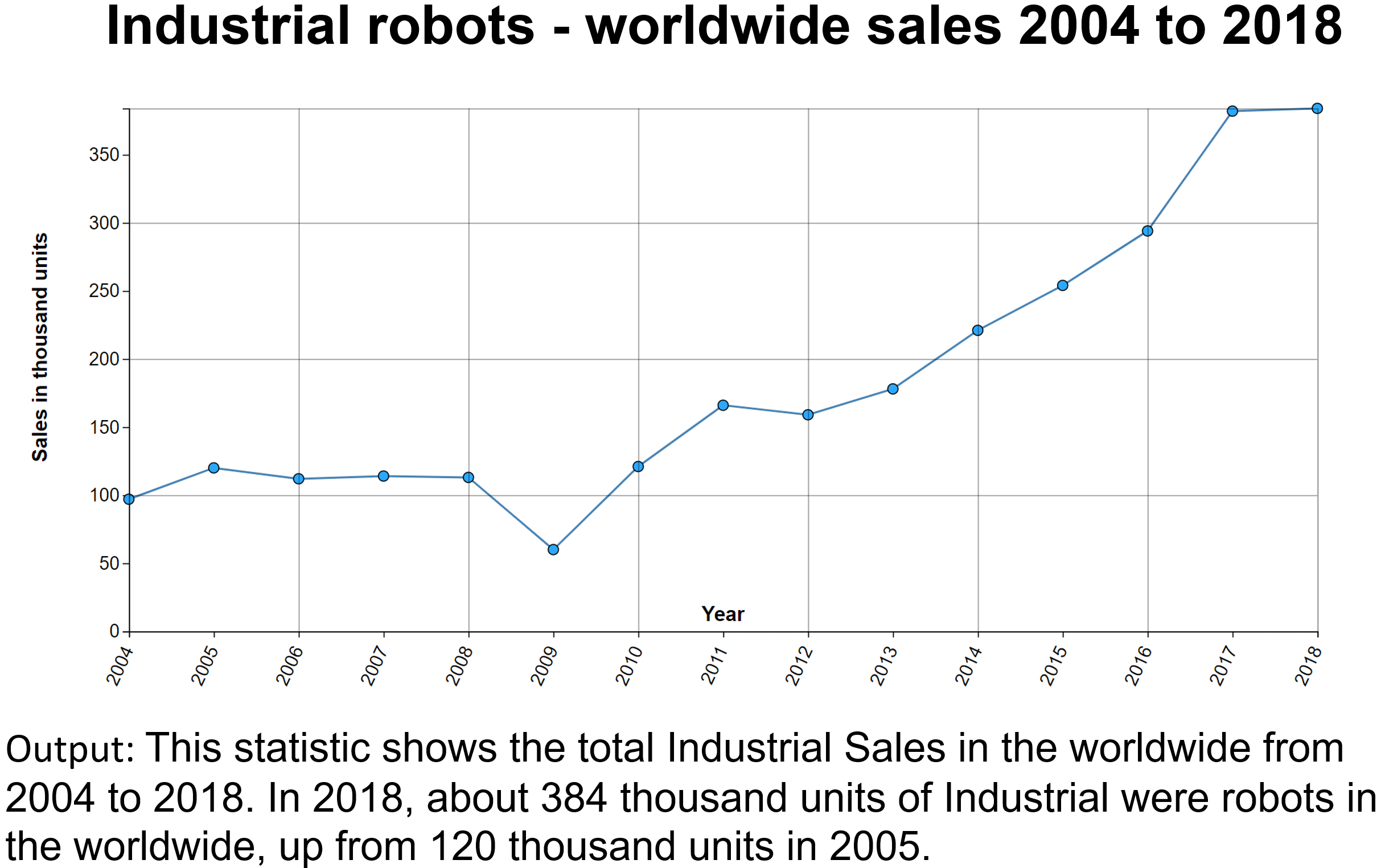}
							\caption{ Example outputs of an automatic chart summarization model. The textual summary attempts to explain important insights from the chart such as trends and extreme values \cite{charttotext}.  
							} 
							\label{fig:charttotext}
						\end{figure}

						\begin{figure}[t!]
							\includegraphics[width=0.9\linewidth]{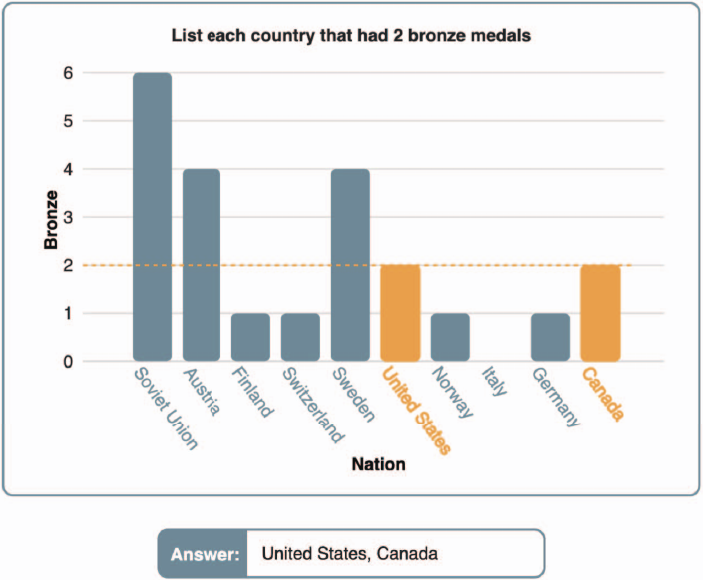}
							\centering
							\caption{An example of visualization as output from the Advisor system \cite{advisor}. 
								The query \textit{``List each country that had 2 bronze medals?''} is asked regarding a data table. In response, the system returns the textual answer in the \textit{``Answer:''} bar besides providing a visualization (bar chart) in which the number of bronze medals of each country is exhibited and the two countries that are answers of the query are highlighted \cite{advisor}}
							\label{fig:advisor}
						\end{figure}

						\section{Output Dimension}
						\label{sec:output}
						As shown in Figure~\ref{fig:introduction}, similar to Input Dimension, outputs in CQA are shown in \textit{Text}, \textit{Visualization}, or \textit{Multimodal}. We discuss each of these categories in detail.
						
						\subsection{Textual}
						\label{sec:input-textual}
						Most CQA models focus on answering questions that require a textual answer. Usually,  such an answer is either a \textbf{numeric token} or a \textbf{word/phrase} 
						\cite{
							leafnet, dvqa, figureqa, prefil}. Such textual outputs can be categorized  into \textbf{fixed vocabulary} vs. \textbf{open vocabulary}. In case of the fixed vocabulary output setting, the answer comes from a small fixed-sized vocabulary (e.g. chart axis labels, bar values, `yes', `no'). For example, early systems  outputs only \texttt{yes} or \texttt{no} to the questions about the chart \cite{figurenet, figureqa}. Some systems support other possible output textual elements in chart and certain numeric values as they add them to the fixed vocabulary \cite{dvqa, prefil, leafnet, stlcqa, affinity}. These models that only support \emph{fixed vocabulary} questions usually treat the task as a classification problem and rely on dynamic encoding techniques where the questions and answers are encoded in terms of spatial positions of chart elements (e.g. \textit{x-axis-label-1}, \textit{x-axis-label-2} and so on). Such approaches do not produce correct outputs when the OCR model for extracting text from a chart generates errors or when the question refers to chart elements using synonyms (e.g., \texttt{US} vs. \texttt{United States}). Overall, treating the CQA problem as a classification task with a fixed vocabulary output is very limited for practical purposes, as it ignores many complex reasoning questions where the answer is derived through various mathematical operations such as aggregation and comparison.
						
						Open vocabulary questions could be more challenging as the system  no longer treats the problem as a classification task and instead needs to derive the answer through analytical operations. However, the works focusing on generating open vocabulary outputs are still very limited~\cite{kim2020, plotqa, ahmed-workshop-2021, evizeon}. Most of these works apply a table question answering model named Sempre~\cite{sempre}. However, Kim et al. simplify the problem by assuming that the underlying data table is available~\cite{kim2020} where others try to extract the data  using computer vision techniques \cite{plotqa, ahmed-workshop-2021}. One limitation with PlotQA is that after automatically extracting the data table it directly applies the Sempre~\cite{sempre}  while answering questions which does not consider any visual features of a chart. As a result their model accuracy can be low for visual questions that makes references to graphical marks and their attributes in chart.

						Moving beyond outputting a word/phrase, an open-ended question that requires  explanatory sentences or paragraphs has not been supported by existing works. For example, to answer the question, \textit{``How has the GDP of Brazil changed over time?''} from the line chart Figure~\ref{fig:chartplots} (d), the output needs to be an explanatory paragraph summarizing the major trends. Recent works on automatic caption generation from charts (e.g., \change{\cite{charttotext, chart-to-text-acl}} can serve as a starting point for generating explanatory answers (see Figure~\ref{fig:charttotext}). Another way a descriptive sentence/paragraph might be helpful is by explaining how the model computes the answer to enhance the transparency of the model to the user. Kim et al. take the intermediate logical form representation of the table question answering system and then translate it into an explanation using pre-defined templates by applying a rule-based approach \cite{kim2020}. Overall, generating explanatory answers to a question about charts remains extremely under-explored.

						\begin{figure}[t!]
							\includegraphics[width=.48\textwidth]{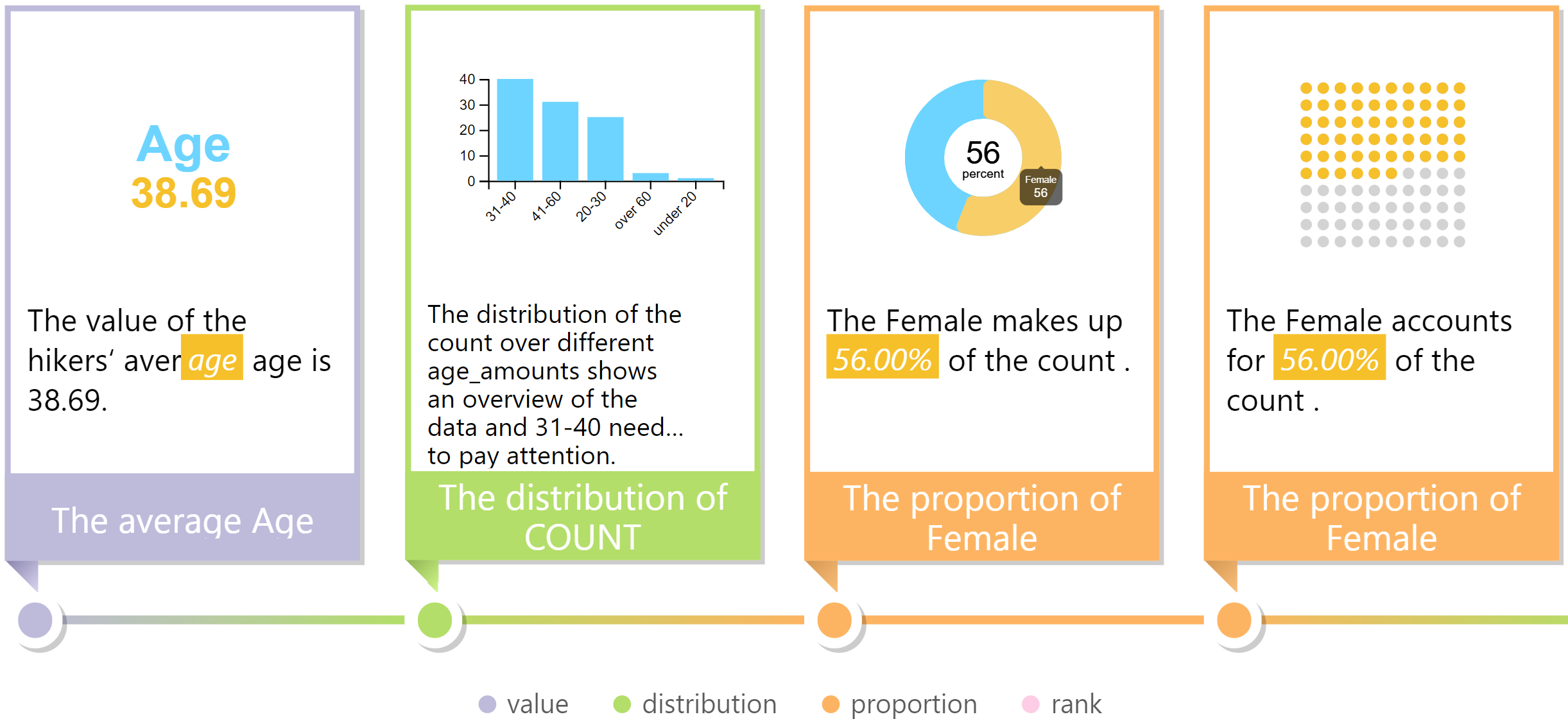}
							\centering
							\caption{ An example output from the Calliope system which automatically generates a visual data story as a sequence of visualizations and texts\cite{shi2020calliope}. Here the visual story summarizes data about hiking using a combination of multiple different visualizations including  bar and pie charts.  
							}
							\label{fig:storytelling}
						\end{figure}
						
						\subsection{Visualization} Some natural language interfaces for visualizations 
						generate an answer by creating a new visualization ~\cite{datatone, advisor, nl4dv} based on the given question and a data table (see Figure \ref{fig:advisor}). Others considered that a visualization is already given as input and the goal is to highlight answers within that existing visualization~\cite{eviza, orko, sneak-peak} (see the example in Figure \ref{subfig:outputvisa}). In an early work, DataTone generates simple charts to answer questions about data tables using dependency parsing and a rule based approach~\cite{datatone}. Evizeon demonstrates a method for adding multiple visualizations using a hand-crafted grammar-based parsing approach. NL4DV \cite{nl4dv} is a Python toolkit that can take a question and a table as input and then output related data attributes, analytical tasks and Vega-Lite specifications. This enables  developers that may not be experts in natural language processing to automatically create the desired visualization using the specifications in the system output. The toolkit interprets the query using a combination of dependency parsing and a rule-based approach. 
						
						The above body of work  generally relies on some heuristics which may put restrictions on possible user input, therefore more robust methods are needed to interpret natural language with rich syntactic and semantic structure. 
						\change{In this direction, Luo et al. \cite{luo2021natural} utilize a transformer-based sequence-to-sequence (seq2seq) model that  translates a natural language statement to visualization.}
						Another system named ADVISor \cite{advisor} attempts to address the problem by applying a deep learning model that used pre-trained Bidirectional Encoder Representations from Transformers (BERT) model to encode the question and the table header and then feeding them to two fully-connected neural networks in order to determine the aggregation operation and related attributes. Finally, the model decides which type of visualization to show based on the aggregation operation and required attribute types extracted from the tabular data and the user's question. Figure \ref{fig:advisor} shows an example output of this approach. 
						
						One aspect that has been rarely covered in terms of outputting visualizations in CQA is presenting multiple views as an answer. Providing different views from the same dataset may help users to grasp different perspectives of the data \cite{multipleviews}. 
						\change{In this regard, combining automatic visualization recommendation \cite{zhu2020survey, zeng2021evaluation, wongsuphasawat2015voyager, vij2022vizai} with chart question answering can be a promising direction.} 
						Future work may explore how to generate multiple views as an answer to provide more perspective perhaps by incorporating visualization recommendation algorithms within the CQA system, especially when the question is open-ended in nature.

						
						\begin{figure}[t!]
							\includegraphics[width=0.5\textwidth]{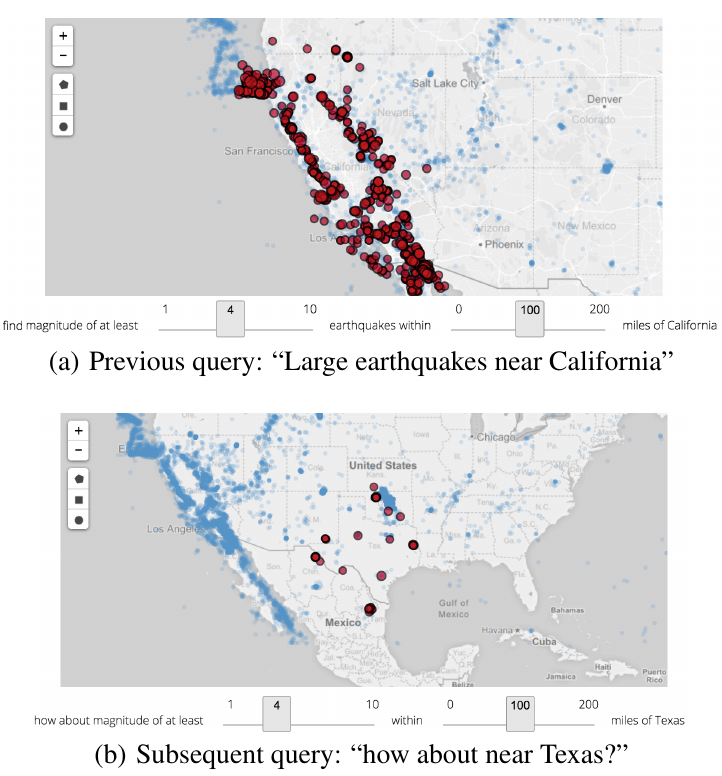}
							\centering
							\caption{An example of an analytical conversation regarding a visualization \cite{eviza}.}
							\label{fig:conversation}
						\end{figure}
						
						\subsection{Multimodal}
						A multimedia output that leverages the complementary power of text and visualization could facilitate users to comprehend the answer more effectively than either visualizations or texts. Given the question \textit{``How have the house prices in Toronto changed over time?''}, a line chart could show the average house price over time while the text could describe the price trends. However, this problem scenario has been virtually unstudied.
						For example,  
						Orko~\cite{orko} provides audio  feedback respectively while responding to a query. But none of the existing CQA approaches explain the important patterns, trends, or outliers with respect to a query.
						
						The multimodal output of a CQA system could also be presented as a sequence of scenes consisting of several visualizations and texts, thus resulting in a narrative story. There have been some recent efforts on automatically generating visual stories consisting of texts and visualizations~\cite{shi2020calliope, chen2019towards, cui2019text, wang2019datashot}. \change{For example, Calliope explores the data space given by the input spreadsheet to generate data facts and organize them in a story sequence~\cite{shi2020calliope} (see Figure \ref{fig:storytelling}).} Some works took images or charts as input and produce visual stories and interactive charts to elaborate the input in another perspective \cite{tang2020plotthread, zhao2021chartstory, wongsuphasawat2015voyager}. There were also papers focusing on tabular data and spreadsheets \cite{shi2020calliope, wang2019datashot} or text data \cite{metoyer2018coupling, hullman2013contextifier, cui2019text, fulda2015timelinecurator, gove2021automatic} as input in timeline or story generation models.
						For instance, Cui et al. \cite{cui2019text} designed a system that takes a simple factual statement with numeric data and outputs various possible infographics to illustrate the statement visually. However, these works do not take any questions as part of the input. \change{While these works are not solving the chart question answering they can provide us with future directions on how to answer questions by combining visualizations with explanatory texts, which has not been investigated yet.} 
						


						\subsection{Conversation vs. Single QA pairs}
						\label{sec:conv}
						People have tendencies in  engaging in conversations involving a series of related question-answer pairs instead of a single question-answer pair. While the conversational question answering task has been explored in the domain of text data \cite{Reddy2018, qu2019bert} as well as image data~\cite{das2017visual, massiceti2018flipdial}, it has not been explored deeply for chart question answering.  Eviza took a first step towards supporting follow-up conversation~\cite{eviza}, while Evizeon models pragmatic behaviour by executing the current query in the context of past interactions~\cite{evizeon}.  Figure \ref{fig:conversation} shows an example of a conversation in Eviza where the user asks a query \textit{``Large earthquakes near California''} followed by \textit{``how about near Texas?''}. In response, the system shows the large earthquakes near Texas based on the context from the last query. Orko \cite{orko} also attempted to handle follow-up questions regarding a visualization by keeping track of context through a conversational centering approach. For instance, to handle two follow-up questions \textit{``Show the strikes of Real Madrid.''} and \textit{``Now show the defenders.''}, the model keeps \textit{``Real Madrid''} as the center to use it in the following questions. Heart and Tory \cite{hearst2019would} conducted user studies to evaluate the influence of showing charts and graphs in the context of a conversational interface. They found that most of the people who wanted to see charts as part of analytical conversation  preferred to see additional supporting context beyond the direct answer to the question. 
						
						Overall, the conversational question answering is a promising direction as existing works demonstrate the benefits of enabling pragmatics in improving analytical workflow while exploring visualization. However, these investigations are preliminary and have not been tested on any benchmark datasets. In future, it is necessary to build large-scale benchmark datasets to systematically evaluate different methods using both quantitative and qualitative measures. Moreover, advanced sequence-to-sequence deep learning method may capture the complex pragmatic aspects of the conversational interfaces more effectively than simple conversational centering~\cite{evizeon, orko} or finite state machine based approach~\cite{eviza}. 
						
						\subsection{Discussion}
						
						\change{In this section, we identified several key limitations and future opportunities along the output dimensions. First of all, many CQA systems are limited to outputting textual tokens with a fixed vocabulary. Future work may explore how to effectively solve open-vocabulary questions and open-ended questions that require explanatory answers. Second, using both texts and visualizations together as the output of the CQA model could contribute to the comprehensiveness and transparency of the answers for users' queries. Third, answering a question by generating multiple views or by generating a visual story can be an interesting direction. Section \ref{sec:challenges} explains more details about the possible challenges and directions for the output dimension of CQA systems.
							
						}

					\section{Evaluation}
					\label{sec:evaluation}
					
					\begin{table*}[]
						\setlength\extrarowheight{2pt}
						\centering
						\caption{
							Comparisons among some existing datasets \change{for chart question answering}. The top five datasets contain automatically generated questions through template-based approach and were developed for the purpose of training and evaluating CQA models. \change{The last two rows introduce datasets with human-written questions.
								%
						}}
						\label{my-label}
						
						\scalebox{0.7}{\begin{tabular}{|l|l|l|l|l|l|l|l|l|l|l|l|l|}
								
								\cline{1-13}
								
								\multirow{4}{*}{} & \multicolumn{11}{c|}{\change{Dimensions}} &   \\ \cline{2-12} 
								
								& \multicolumn{8}{c|}{\change{Input}} & \multicolumn{3}{c|}{\change{Output}} &  \\ \cline{2-12}
								
								& \multicolumn{4}{c|}{\change{Text}} & \multicolumn{4}{c|}{\change{Visualization}} &  \multicolumn{2}{c|}{\change{Text}} & \change{Visualization} & \multirow{3}{*}{\change{\makecell{\#Charts/\\\#QA pairs}}} \\ 
								\cline{2-9}
								\cline{10-12}

								\change{Related Papers} & \multirow{2}{*}{\change{\makecell{Factual/\\Open-ended}}} & \multirow{2}{*}{\change{\makecell{Visual/\\Non-visual}}} & \multirow{2}{*}{\change{\makecell{Simple/\\Compositional}}} &
								\multirow{2}{*}{\change{\makecell{Question\\Types}}} &
								\multirow{2}{*}{\change{\makecell{Bitmap/\\Data Table}}} & \multirow{2}{*}{\change{\makecell{Real-world\\Charts}}} & \multirow{2}{*}{\change{\makecell{\#Chart\\Types}}} & \multirow{2}{*}{\change{\makecell{Real-world\\Data}}} & \change{\makecell{Fixed\\Vocab}} & \change{\makecell{Open\\Vocab}} & \multirow{2}{*}{\change{\makecell{New/\\Existing}}} & \\ 
								
								\cline{10-11} 
								
								& & & & & & & &  & \change{\makecell{Numeric/\\Word}} & \change{\makecell{Numeric/\\Word/\\Sentence}} & & \\
								
								\hline
								
								\rowcolor{gray!30}
								\change{LEAFQA\cite{leafnet}} & \cmark/\xmark & \cmark/\cmark & \cmark/\cmark & \change{\makecell{Template\\Based}} & \cmark/\xmark & \xmark & 6 & \cmark & \cmark/\cmark & \xmark/\xmark/\xmark & \xmark/\xmark & \change{240K/2M}\\ \cline{2-13}
								
								\change{LEAFQA++\cite{stlcqa}} & \cmark/\xmark & \cmark/\cmark & \cmark/\cmark & \change{\makecell{Template\\Based}} & \cmark/\xmark & \xmark & 6 & \cmark & \cmark/\cmark & \xmark/\xmark/\xmark & \xmark/\xmark & \change{244K/2.5M} \\ \cline{2-13} 
								
								\rowcolor{gray!30} 
								\change{DVQA\cite{dvqa}} & \cmark/\xmark & \cmark/\cmark & \cmark/\cmark & \change{\makecell{Template\\Based}} &  \cmark/\cmark & \xmark & 1 & \xmark & \cmark/\cmark & \xmark/\xmark/\xmark & \xmark/\xmark & \change{300K/3.4M}  \\ \cline{2-13}

								\change{PlotQA \cite{plotqa}} & \cmark/\xmark & \cmark/\cmark & \cmark/\cmark  & \change{\makecell{Template\\Based}} & \cmark/\xmark & \xmark & 3 & \cmark  & \cmark/\cmark & \cmark/\cmark/\xmark & \xmark/\xmark  & \change{224K/28M}  \\ \cline{2-13}
								
								\rowcolor{gray!30} 
								\change{FigureQA\cite{figureqa}} & \cmark/\xmark & \cmark/\cmark & \cmark/\cmark & \change{\makecell{Template\\Based}} & \cmark/\cmark & \xmark & 4 & \xmark  & \cmark/\cmark & \xmark/\xmark/\xmark & \xmark/\xmark  & \change{180K/2.3M}  \\ \cline{2-13}
								
								\specialrule{.8pt}{0pt}{0pt}%
								
								\change{Kim et al.\cite{kim2020}} & \cmark/\xmark & \cmark/\cmark & \cmark/\cmark  & \change{\makecell{Human\\Authored}} & \xmark/\cmark & \cmark & 2 & \cmark & \cmark/\cmark & \cmark/\cmark/\cmark & \xmark/\xmark & \change{52/629} \\ \cline{2-13}
								
								\rowcolor{gray!30} 
								
								\change{Masry et al.\cite{Masry2022ChartQAAB}} & \cmark/\xmark & \cmark/\cmark & \cmark/\cmark & \change{\makecell{Human\\Authored}} & \cmark/\cmark & \cmark & 3 & \cmark & \cmark/\cmark &\cmark/\cmark/\xmark & \xmark/\xmark  & \change{4.8K/9.6K}  \\ \cline{2-13}
								\hline
								
						\end{tabular}}
						\label{tab:datasets}
					\end{table*}
					
					\changesecond{In this section, we discuss the various evaluation techniques for chart question answering. We group the evaluation methods used in the surveyed papers based on the major dimensions of the CQA problem space: (1) Input Dimension, (2) Output Dimension. 
						
					}
					

					\begin{figure*}[h]
						\begin{subfigure}[b]{0.24\textwidth}
							\centering
							\scalebox{0.94}{\begin{tabular}{l}
									{\includegraphics[width=\linewidth]{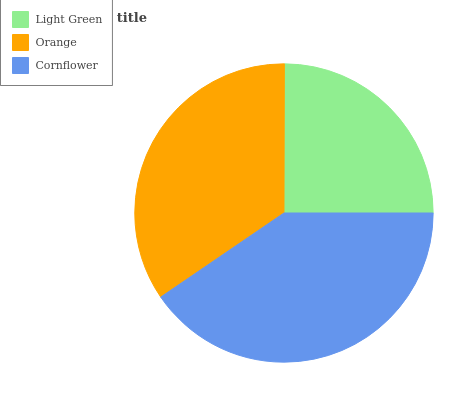}}
									\\
									{\begin{tabular}{l}
											{\tiny{\textbf{Q:} \tiny Is Orange less than Light Green?}}
											\\
											{\tiny{\textbf{A:} No}}
										\end{tabular}
									}
								\end{tabular}
							}
							\caption{FigureQA Pie Chart \cite{figureqa}}
							\label{subfig:figureqa_example}
						\end{subfigure}
						\begin{subfigure}[b]{0.24\textwidth}
							\centering
							\scalebox{0.94}{\begin{tabular}{l}
									{\includegraphics[width=\linewidth]{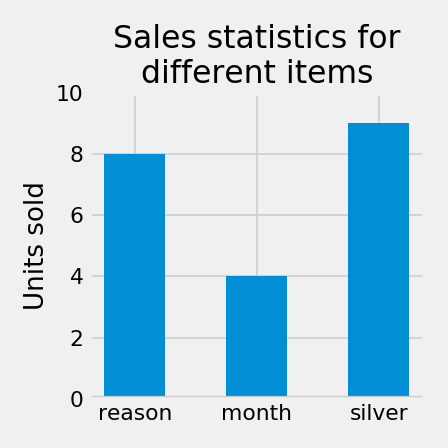}}
									\\
									{\begin{tabular}{l}
											{\tiny{\textbf{Q:} \tiny Did the item month sold less units than reason?}}
											\\
											{\tiny{{\textbf{A:} Yes}}}
										\end{tabular}
									}
								\end{tabular}
							}
							\caption{DVQA Bar Chart \cite{dvqa}}
							\label{subfig:dvqa_example}
						\end{subfigure}
						\begin{subfigure}[b]{.24\textwidth}
							\centering
							\scalebox{0.94}{\begin{tabular}{l}
									\raisebox{-0.5\height}{\includegraphics[width=\linewidth]{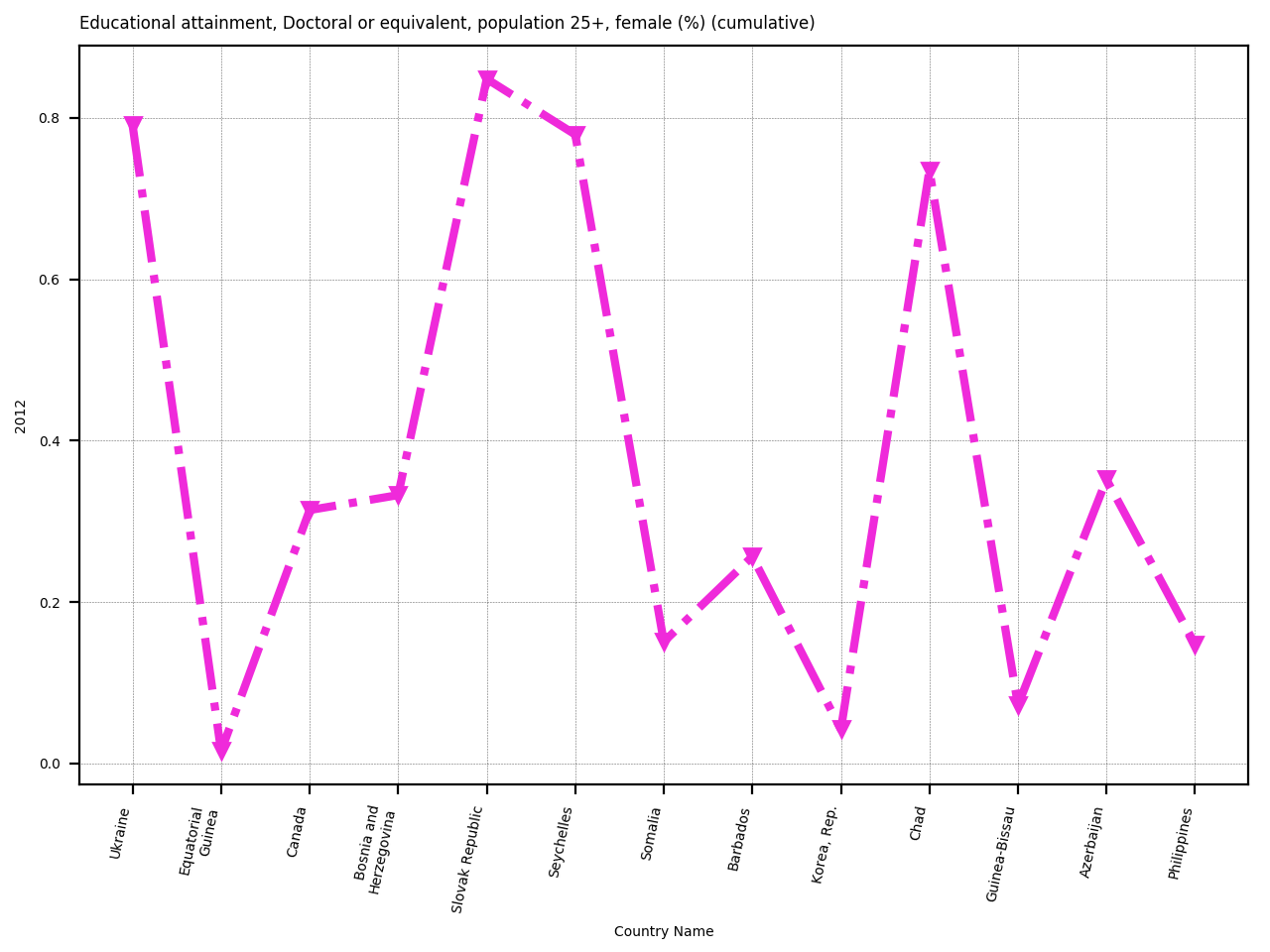}}
									\\
									{\begin{tabular}{@{}p{3.9cm}@{}}
											{\tiny{\textbf{Q:} \tiny Indicate the Country Name that has the Educational attainment, Doctoral or equivalent, population 25+, female (\%) (cumulative) of 2012 less than the year Guinea-Bissau?}}
											\\
											{\tiny{\textbf{A:} Korea, Rep.}}
										\end{tabular}
									}
								\end{tabular}
							}
							\caption{LeafQA Line Chart \cite{leafnet}}
							\label{subfig:leafqa_example}
						\end{subfigure}
						\begin{subfigure}[b]{.24\textwidth}
							\centering
							\scalebox{0.94}{\begin{tabular}{l}
									\raisebox{-0.5\height}{\includegraphics[width=\linewidth]{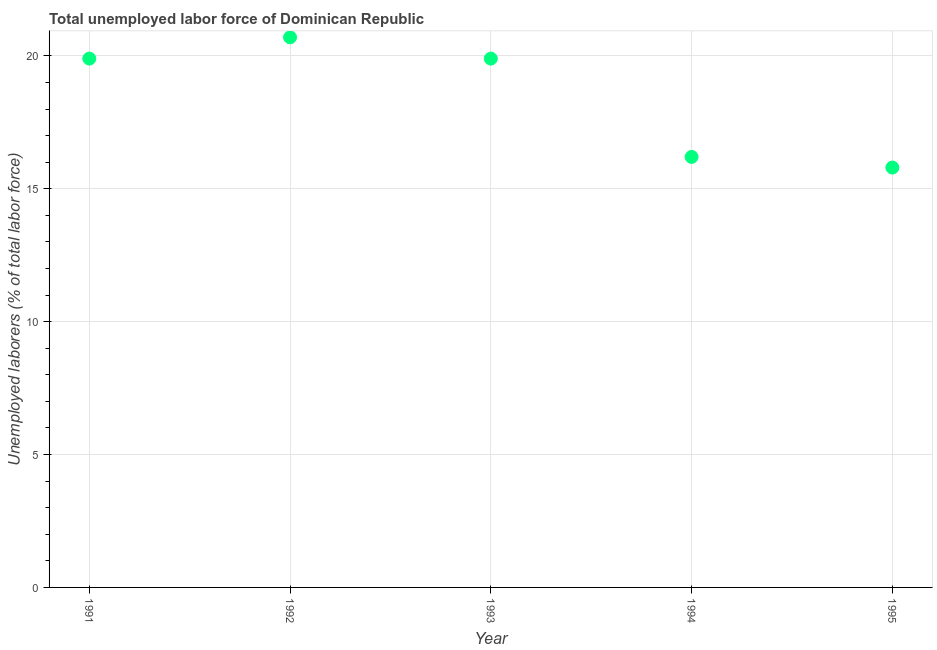}}
									\\
									{\begin{tabular}{@{}p{3.9cm}@{}}
											{\tiny{\textbf{Q:} \tiny  What is the difference between the total unemployed labour force in 1993 and 1994 ?}}
											\\
											{\tiny{\textbf{A:} 3.7}}
										\end{tabular}
									}
								\end{tabular}
							}
							\caption{PlotQA dot line chart \cite{plotqa}}
							\label{subfig:plotqa_example}
						\end{subfigure}
						\caption{Some example charts from existing CQA datasets along with some of their QA pairs. Due to using randomly-generated data, the questions in \ref{subfig:figureqa_example} and \ref{subfig:dvqa_example} lack naturalness. \ref{subfig:leafqa_example} shows a line chart from LeafQA \cite{leafnet} where the x-axis consists of categorical variables that gives incorrect sense of value change. \ref{subfig:plotqa_example} shows a dot line chart from the PlotQA \cite{plotqa} dataset with open-vocabulary question.  
						}
						\label{fig:example_charts}
					\end{figure*}
					
				
				\subsection{\changesecond{Input Dimension}}
				\label{sec:benchmark}
				
				\subsubsection{\changesecond{Text}}
				
				\changesecond{In terms of the text as input dimension, most benchmarks contain questions about charts that are created using pre-defined templates (see Table~\ref{tab:datasets}). For example, earlier CQA datasets, namely FigureQA \cite{figureqa} and DVQA \cite{dvqa}, consisted of questions generated from a small number of templates. Such} template-based questions lack the lexical/syntactic variations and nuances/erratum that are prevalent in the human-written questions. For example, the question in Figure \ref{subfig:figureqa_example} can be written as `\textit{`Is the area colored in orange smaller than the green area?''}. Although LeafQA \cite{leafnet} and LeafQA++ \cite{stlcqa} applied paraphrasing tools to introduce lexical variations into their questions, they still do not stand on par with the human-authored questions. PlotQA attempted to alleviate the issue through  by gathering a limited number of human-authored questions through crowdsourcing to create more realistic templates for the questions in their datasets~\cite{plotqa}. 
				
				
				
				
				
				Lack of real-world human-annotated questions remains a key bottleneck in solving the chart question answering question. Kim et al.\cite{kim2020} ran a formative study with a very small human-authored dataset consisting of 52 charts and 629 QA pairs to understand how people ask questions about charts and explain answers. Srinivasan et al. \cite{srinivasan2021collecting} runs another study to curate a dataset of 893 utterances that are used to  generate visualizations, out of which 114 were questions while the remaining ones include natural language commands and web search-like phrasal queries.
				However, both of these datasets are quite small and were curated for understanding how people interact with visualizations through natural language rather than utilizing for training models. \change{Masry et al. \cite{Masry2022ChartQAAB} attempted to addressed the limitation by introducing a large number of human written questions (9.6K). Since collecting human-authored questions is time-consuming and costly, they have augmented their dataset with 23.1K machine-generated questions from the Statista chart human-written summaries using the T5 model \cite{Raffel2019}. While such data augmentation through machine-generated question is promising, they are also limited in various ways (e.g, unanswerable questions, lack of visual reasoning questions that refer to chart elements).
					

					
					
					\subsubsection{\changesecond{Visualization}}
					\changesecond{When we consider visualization as input, most benchmark datasets constructed charts synthetically using randomly generated data. Among them,} FigureQA \cite{figureqa} covers five different \textbf{\changesecond{chart types}} (e.g., bar, line, and pie) while 
					DVQA \cite{dvqa} contains only bar charts.  \changesecond{All benchmarks except \cite{kim2020} contain \textbf{bitmap images} of charts as input visualizations. FigureQA \cite{figureqa}, DVQA \cite{dvqa}, and Masry et al. \cite{Masry2022ChartQAAB} provide \textbf{underlying data} of input charts alongside the bitmap images. Kim et al \cite{kim2020} only consider underlying data table as the input chart to CQA model.} To generate the charts' textual labels (e.g., \textit{x-axis} labels) synthetically, FigureQA utilized the X11 color set and DVQA \cite{dvqa} used the most common 1000 vocabulary from the Brown Corpus. While such approaches can help to develop a large dataset automatically, the randomly generated data and textual labels caused the charts and the questions to lack realism and meaningfulness. For example, Figure \ref{subfig:dvqa_example} shows a bar chart in the DVQA dataset with three completely unrelated categorical labels (`silver', `month', and `reason') along with the corresponding question  \textit{"Did the item month sold less units than reason?"}, which sounds unnatural and semantically meaningless. LeafQA \cite{leafnet} and LeafQA++ \cite{stlcqa} datasets entail new chart types such as box and scatter plots. Unlike the previous datasets, they utilized real-world tabular data extracted from online sources to synthetically plot their chart images.
					
					Overall, most of the above datasets suffer from several common issues due to their synthetic nature. First, synthetically plotted charts lack the diversity in visual styles and charts' structure that are available in real-world charts. Although some works like PlotQA tried to randomly set visual parameters (e.g., font size, presence of grids), the charts visual styles were still limited due to using only one software tool, namely  Matplotlib~\footnote{https://matplotlib.org/}. Moreover, data attributes were plotted on charts without considering their types, resulting in poorly designed charts. For example, it is well known that plotting categorical data on a line chart is a bad idea~\cite{munzner2014visualization}. However, Figure \ref{subfig:leafqa_example} shows a line chart from LeafQA with categorical attributes and gives a false sense of perceptual change of values from one country to another.
					
					\change{Very recently, Masry et al. \cite{Masry2022ChartQAAB} constructed the ChartQA dataset, consisting of 4.8K real-world charts. 
						To ensure variations in the visual styles, they crawled chart images from four different online sources. 
						Still, their dataset is relatively small compared to the previous large-scale synthetic datasets and only covers three different chart types. Therefore, there is a pressing need for benchmarks covering a large collection of charts with diverse styles. 
						
						\subsubsection{\changesecond{Multimodal}}
						While all the benchmark CQA datasets take only textual query as input, research works related to more broad topics about natural language interfaces for visualizations conducted user studies to evaluate multimodal input features (e.g. \cite{orko, eviza, srinivasan2020inchorus}). In these approaches, experimenters ask participants to have interactions with the designed system (or multiple systems in case of comparison). By analyzing some quantitative measurements (e.g., task completion time) and subjective feedback from participants through questionnaires, interviews regarding their interactions, experimenters gained both quantitative and qualitative insights about the system. Eviza~\cite{eviza} followed two goals of (1) observing qualitative feedback, and (2) comparing NLI and direct manipulation (DM). For qualitative measurement, they followed the think-aloud protocol where participants explained their thoughts during the tasks. To compare natural language interface with a direct manipulation-based interface, they designed the tasks on both their system and Tableau Desktop. Orko~\cite{orko} did the same about qualitative measurement by asking participants to express their thoughts. They also evaluated the multimodal interaction in their system. Orko~\cite{orko} and Inchorus~\cite{srinivasan2020inchorus} used the  \textit{jeopardy evaluation} method which allows experimenters to give participants some \textit{facts} instead of explicit queries. In this way, participants have to interact with the system in a way that they can show the fact by visualization output. Finally, participants were given a questionnaire to tell their opinions and score the system based on different measurements. 
						
						In general, many of these research works are limited in that their systems are not compared with their counterparts, perhaps because many of these NLI systems are not publicly available. Furthermore, most studies have been based on qualitative measurements depending on participants' subjective feedback without much objective quantitative measures. Finally, user studies often took place informally in laboratories by asking a limited number of participants to use the target system which lack realism~\cite{carpendale2008evaluating, lam2011empirical}.

						



						}
						
						\subsection{\changesecond{Output Dimension}}
						\subsubsection{\changesecond{Text}}
						
						In all existing benchmarks, outputs are limited to texts only and in most datasets possible answers are also limited  (Table~\ref{tab:datasets}). \textbf{\changesecond{Fixed-vocab}} datasets such as FigureQA, DVQA, LeafQA and LeafQA++ have a limited number of possible answers that often refer to the charts' textual elements or common answers  such as `yes', `no',  and `all' (see Figure \ref{subfig:dvqa_example}, \ref{subfig:figureqa_example}, and \ref{subfig:leafqa_example}). For example, FigureQA only supports `yes'/`no' answers. DVQA has 25 different templates with 1576 distinct possible answers while LeafQA \cite{leafnet} and LeafQA++ \cite{stlcqa} had 75 possible answers, which mainly consist of textual labels in charts (see Figure \ref{subfig:leafqa_example}) or common answers. Such datasets are limited in that they do not consider questions where the answer cannot be found directly in the chart and instead it is required to be computed through operations such as sum, average, count etc. 
						
						To address the limitation,} PlotQA \cite{plotqa} introduces \textbf{\changesecond{open-vocabulary}} questions that require an answer that can be obtained by aggregating over the underlying data values of the charts (see Figure \ref{subfig:plotqa_example}). \changesecond{
						Still, all the existing CQA datasets have factual \textbf{factual} questions (e.g. Figure \ref{subfig:plotqa_example}). None of the benchmarks contains \textbf{open-ended} questions that require textual explanatory answers. 
					} 
					
				}
				
				\changesecond{\subsubsection{Visualization}
					Some NLIs which generate an answer by creating a new visualization ~\cite{datatone, advisor, nl4dv} mainly focused on demonstrating the capability of the method through analyzing example outputs. NL4DV demonstrated it's capabilities through several application scenarios (e.g., create visualizations
					in Jupyter notebooks) while  AdVISor was compared with NL4DV for various sample queries over a dataset. DataTone~\cite{datatone} was evaluated through a user study following the \textit{jeopardy evaluation}, where it was compared with IBM's Watson. However, as the authors acknowledged that directly comparing these two interfaces was difficult as the two systems looked very different. Overall, a key weakness of evaluating the above system is that they were not evaluated on any benchmark dataset by comparing with any competing systems.

					\subsubsection{Multimodal}
					
					As we have discussed in Section 4.3, existing CQA systems do not produce multimodal output combining explanatory text and visualizations as an answer. A starting point could be to build a benchmark dataset to facilitate development and evaluation of new CQA systems that generate multimodal output.  Since a multimodal output can be presented as a data story (possibly as a sequence of scenes) it is also worth visiting the evaluation techniques for data driven storytelling. Amini et al. identified several key criteria for evaluating data driven stories~\cite{amini2018evaluating}, such as comprehension,  memorability, and engagement which can be useful in the context of answering questions via data-driven stories. If the output is interactive, quantitative metrics such as time spent by users on explanatory story consumption and interaction statistics (e.g., number of clicks)  can give us valuable hints about the efficacy of the generated output. Similarly, self-reported measures such as post-viewing questionnaires and interviews can provide subjective insights about the effectiveness of the generated answer.
				}

					\change{\subsection{Discussion} In this section, we identified the characteristics and limitations of evaluation techniques for CQA systems.} Existing benchmark datasets lack realism in various ways and there is a pressing need for large-scale benchmark datasets consisting of  human-authored questions and real-world charts in solving the chart question answering problem. A starting point could be to utilize several existing real-world large-scale datasets containing visualizations that cover a variety of chart types and diverse visual styles \cite{battle2018beagle,hoque-vis-search-2019, chen2021vis30k, deng2020visimages, chen2020composition}. For example, Beagle is a dataset with 41K real-world charts crawled from the Web  that were created using five different tools\cite{battle2018beagle}. Hoque and Agrawala
					\cite{hoque-vis-search-2019} built a  search engine for visualizations by crawling a collection of 7.8K D3 charts from the Web and deconstructing each one to recover its data. Others have collected visualizations and figures from scientific papers~\cite{siegel2016figureseer, chen2020composition, deng2020visimages, chen2021vis30k}. \change{ Moreover, we may consider converting existing TableQA datasets \cite{sempre, zhong2017seq2sql, SQA} into ChartQA datasets by translating their queries to SQL commands \cite{text2sql} and plotting the relevant portions of the data tables into chart images using SQL2Visualization methods \cite{sql2vis1, sql2vis2}. With respect to user studies, }future works need to focus on field trials and longitudinal studies where the participants can ask their own questions with their own datasets. This will help  to understand the utilities, trade-offs and adoption rate of CQA systems in a more realistic way.

					\begin{figure*}[h]
						\includegraphics[width=\textwidth]{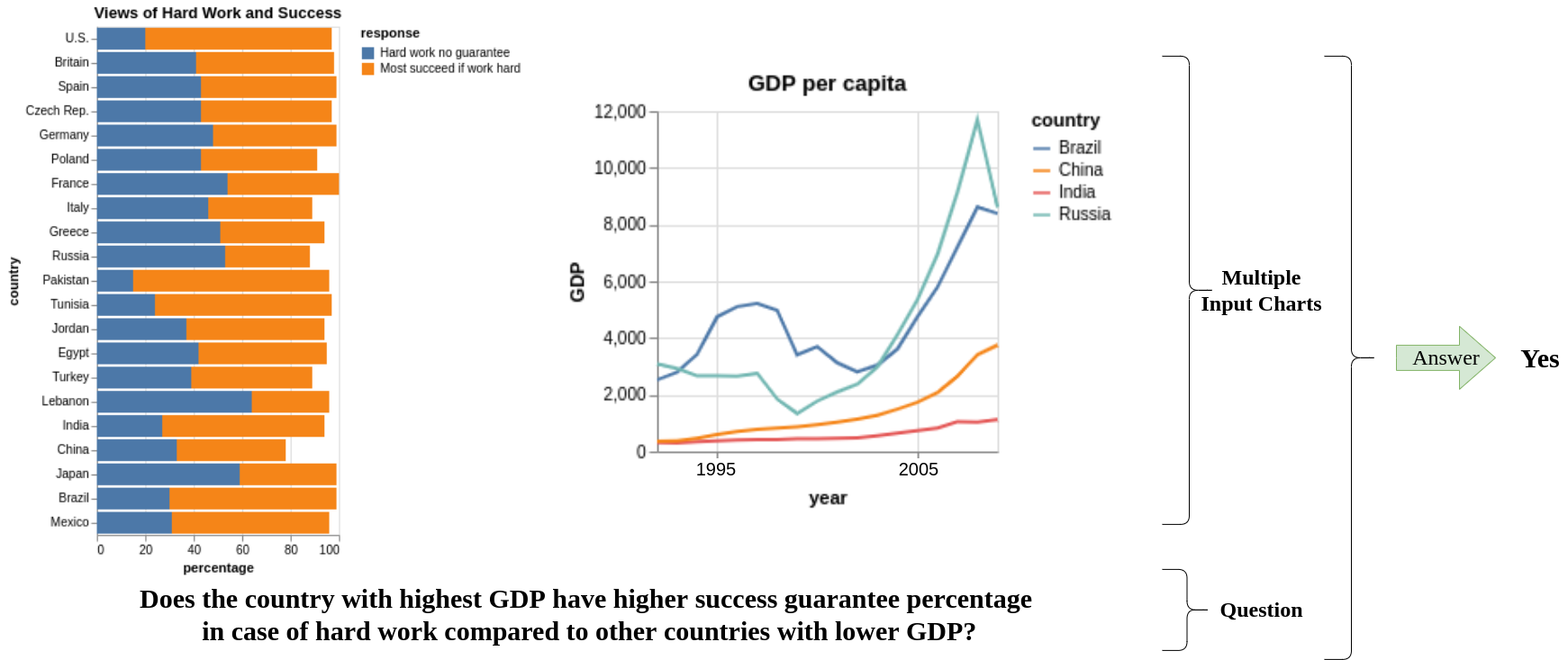}
						\centering
						\caption{An example of multiple charts as input. The charts are from the dataset used in \cite{kim2020}. The question requires information from both charts. First, the information for \textit{``the country with highest GDP''} should be found in the right chart and then, by analyzing the left chart, the \textit{Most succeed if work hard} component of all countries existing in the right chart are compared to each other in order to reach the final answer.}
						\label{fig:multiplecharts}
					\end{figure*}

					\section{Challenges and Research Opportunities}
					\label{sec:challenges}
					
					So far, we have presented a formal taxonomy of the chart question answering task and categorized the existing solutions and evaluation techniques around the CQA problem space we identified. This systematic review has helped us identify the major gaps in current literature and future opportunities in this exciting area at the intersection of information visualization and AI. We now present  these major challenges and possible future directions.
					
					\noindent \textbf{$\bullet$ Constructing more benchmark datasets}: As we have discussed in Section~\ref{sec:benchmark}, most existing CQA datasets mainly provide template-based questions and synthetically plotted chart images which lack realism. To address this key limitation, it is necessary to develop human authored questions and real-world chart sources with a variety of chart types and visual styles. Human-authored questions can pose several challenges for the CQA models due to the rich semantics, language variations, informal languages and typos. 
					Moreover, there are several types of questions from Table~\ref{tab:examples} that are underexplored in the existing datasets. While the PlotQA dataset \cite{plotqa} supported open-vocab questions, they were mostly data-related questions without focusing on questions that refer to the visual attributes (e.g., \textit{color}, \textit{position}) of the chart elements. Finally, there is no existing datasets for open-ended questions that require explanatory paragraphs. Similarly, while there have been some work on conversational interfaces for visualizations~\cite{Revision}, there is no existing dataset that would allow us to train and evaluate conversational question answering interfaces quantitatively. Constructing new datasets on such under-explored problems could open up exciting new avenues of chart question answering.
					
					\noindent \textbf{$\bullet$ Leveraging more advanced deep learning models}}:
				There are significant rooms for improving the existing methods for chart question answering. Many natural language interfaces for charts often rely on simple heuristics that fail to understand questions that are compositional and require a deeper understanding of the syntactic, semantic, and pragmatic aspects \cite{eviza, evizeon, orko, flowsense}. This is a serious limitation, as the user becomes very frustrated if the system frequently fails to understand the query correctly. In contrast, existing works that solely focus on chart question answering apply classification techniques on chart images~\cite{figureqa, dvqa, leafnet} or table question answering on the data table of the  chart~\cite{plotqa, kim2020}. As we discussed in Section~\ref{sec:input-textual}, both  approaches are limited in handling many real-world questions. To address the limitation, future models need to effectively combine both the visual features and the data values of a chart in a unified representation. In this regard, adapting models from the Vision-Language domain\cite{lxmert, vit, VideoBERT, vlt5, albaf}, which unify visual and textual features effectively through transformer-based models can be a starting point. 

				\noindent \textbf{$\bullet$ Addressing chart data extraction challenges}:} 
			Many CQA tasks involving perceptual ~\cite{haehn2018evaluating} and arithmetic reasoning with charts, which require accurate extraction of chart data and visual encodings (i.e., how data is mapped to visual attributes of graphical marks).  Many natural language interfaces assume that the underlying data table and visual encodings of the charts are readily available \cite{eviza, orko, evizeon, sneak-peak}. This assumption becomes invalid for most real-world charts on the Web which are available in bitmap image format without the underlying data. For other charts that are available in SVG format (e.g., D3 charts), there are some techniques for extracting data and visual encodings, however they are also error-prone in some situations (e.g., cannot accurately extract map chart data)~\cite{harper2017converting, hoque-vis-search-2019}.  
			
			In general, chart data extraction from bitmap images is more challenging and requires us to apply computer vision techniques to automatically extract the underlying data. PlotQA~\cite{plotqa} attempts to use automatically extracted data from chart images to perform question answering, however, it suffers from low accuracy due to data extraction errors. The key problem here is that existing solutions for chart data extraction are quite limited. Some of them are not fully automatic~\cite{Revision, jung2017chartsense} which is not very useful for practical purposes. Some focused on recovering color encodings from various charts~\cite{
				poco2017extracting,yuan2021deep}. Others  provided automatic solutions but they rely on various heuristics which do not work for many real-world charts and the performance is still not high enough~\cite{Choi2019VisualizingFT, Liu2019DataEF} . ChartOCR automatically extracts data from real-world charts with reasonably high accuracy~\cite{ChartOCR} but the model only predicts the raw data values of marks (e.g., bars) without associating them with their corresponding axis or legends. Overall, current approaches for  automatic data extraction are usually modular and rely on assumptions which are error-prone. An end-to-end deep learning approach could help improve the performance and generalize well to different chart styles.

			\noindent \textbf{$\bullet$  Answering questions with multiple views}: So far, chart question answering has been explored mainly in the context of a single chart as input while only a few natural language interfaces have demonstrated some interactions with multiple views~\cite{evizeon, sneak-peak}. In reality, people often interactively analyze dashboards and multiple coordinated views. Analytical questions with multiple views can be more challenging as questions can be more compositional, involving references to multiple charts. Figure \ref{fig:multiplecharts} shows an example of a question about multiple input visualizations. As we can see, the question requires information from both charts and the model needs to first retrieve the country with \textit{``Maximum GDP value''} in the related chart (the right one) and then find the \textit{``Success guarantee percentage in case of hard work ''} of that country and other countries in the left chart to make the comparison. In this context, a starting point could be to collection large-scale real-world human-annotated questions involving multiple charts to better understand and characterize the problem.

			\noindent \textbf{$\bullet$ Combining texts and visualizations as answers}}: Existing question answering interfaces for data visualizations typically convey answers in the form of either textual (e.g., \cite{figureqa, dvqa, leafnet, plotqa}) or visual representations (e.g.~\cite{datatone, advisor} \change{\cite{luo2021natural}}) only. \change{Srinivasan et al\cite{srinivasan2018augmenting} provided users with a system whereby they can interact with automatically generated textual data facts to search for possible visualizations.} Automatically generating a multimedia output combining both text and visualization would not only facilitate users to comprehend the results more effectively by explaining key points but also enhance the transparency of the algorithm by conveying how the results were computed. For example, given the question \textit{``How have the house prices in Toronto changed over time?''} a line chart could show the average house price over time while the text could summarize the price trends. While there have been some works for automatically generating chart summary to describe the patterns, trends and outliers in the chart\cite{charttotext}, more effort is needed to generate such text in the context of an open-ended question, where generating a chart and the related explanatory text is very helpful to users. Building on  deep learning models with encoder-decoder architectures from the NLP and Vision-Language domains \cite{lxmert, vit, VideoBERT, vlt5} can be a promising future direction to generate such open-ended explanatory answers.
		
		\noindent \textbf{$\bullet$ Answering questions via visual data storytelling}}: Another future avenue could be combining chart question answering with visual data storytelling to answer the user’s questions. The idea here is that rather than showing all the insights about a high-level question (e.g., \textit{``what is really warming the world?''}) at once, a story creates suspense and makes it easy to comprehend insights by transitioning through a sequence of visualizations \change{~\cite{riche2018data,gershon2001storytelling, kwon2014visjockey, segel2010narrative}}. For instance, given the above question, a manually designed data story went through a sequence of line charts to explain which factors have been argued to contribute to global warming to lead to the conclusion that greenhouse gas is indeed the main factor for global warming~\footnote{https://www.bloomberg.com/graphics/2015-whats-warming-the-world/}. But can we generate such data-driven stories automatically? Recently, there have been some attempts to automatically generate stories from data, however, they do not take any questions into account
	\cite{shi2020calliope, chen2019towards, cui2019text, wang2019datashot}. Generating an interactive data story as a sequence of scenes would require us to answer to questions: \textit{(i)} what to say in the story?  \textit{(ii)} how to say it? Solving this problem is challenging as it would be very difficult for a machine to build a creative narrative automatically that highly professional data journalists build manually. To address this challenge, future work may investigate how the system can automatically build a narrative structure with a sequence of ‘scenes’ combining text and visualizations in a coherent way~\cite{hullman2011visualization, riche2018data}. In future,  machine learning models may learn narrative structures from a large collection of annotated human-authored visual data stories. As a starting point, the model could focus on how to learn a simple linear narrative, then followed by generating complex non-linear narrative structures. \change{Linking texts and visualizations  \cite{zhi2019linking, kim2018facilitating} can be also helpful to readers while answering questions through storytelling.} 
	
	\noindent \textbf{$\bullet$ Leveraging question answering for chart accessibility} Although the field of visualization has grown dramatically in recent years, the research on inclusive and accessible visualization design remains under-explored~\cite{Kim2021}.
	Specific visualization tools such as SAS~\footnote{www.sas.com} and HighCharts~\footnote{https://www.highcharts.com/docs/accessibility/accessibility-module} provide limited support by reading data values from charts based on keyboard interactions which can be difficult and mentally taxing when the number of data points is large. While encountering visualizations on Web, blind users largely depend on screen-reader tools that read the alternative text or caption embedded to the chart. However, such alternative texts and captions are often  not much helpful or not available at all \cite{Morris2018}. A recent study found that chart description should explain important trends and key statistics rather than simply saying how the data is encoded~\cite{accessibile-viz}. 
	In this context, introducing chart question answering could significantly advance the field of accessible data visualizations~\cite{sharif2022voxlens}. In particular, future research could combine automatic chart summarization~\cite{charttotext} with chart question answering so that people who are blind or have impaired vision can use the audio description of a chart and then simply ask various questions about the chart via speech rather than navigating through numerous data points.

	\section{Conclusion}
	
	In this paper, we have presented a survey on chart question answering by analyzing relevant papers from the field of information visualization, human computer interactions, and natural language processing. Through this analysis, we derived a taxonomy with the possible input and output dimensions that illustrate the problem space. We synthesized the findings from existing works along these dimensions and identified key knowledge gaps in this domain. Finally, we outlined the open challenges and opportunities  to inform
	future research in the domain. We hope that this survey will help other researchers in initiating future contributions in this relatively new area of visualization research.

	\section*{Acknowledgement}
	This work was supported by the Natural Sciences and Engineering Research Council (NSERC), Canada. We thank
	anonymous reviewers for their valuable comments and suggestions.

	\bibliographystyle{eg-alpha-doi}
	\bibliography{cqa}
	
	
\end{document}